\documentclass{cta-author}
\usepackage{url}
\usepackage{epstopdf}
\usepackage{multirow}
\usepackage{booktabs}
\usepackage{todonotes}
{}
{}
{}
\def\ie{{\em i.e.}}
\def\eg{{\em e.g.}}
\def\vs{\emph{vs.}}
\def\etal{{\em et al.}}

\begin{document}
\supertitle{Submission Template for IET Research Journal Papers}
\title{Cross-ethnicity Face Anti-spoofing \\ Recognition Challenge: A Review}
\author{
\au{Ajian Liu$^{1}$},
\au{Xuan Li$^{2}$},
\au{Jun Wan$^{3\corr}$},
\au{Yanyan Liang$^{1}$},
\au{Sergio Escalera$^{4}$},
\au{Hugo Jair Escalante$^{5}$},
\au{Meysam Madadi$^{4}$},
\au{Yi Jin$^{2}$},
\au{Zhuoyuan Wu$^{6}$},
\au{Xiaogang Yu$^{6}$},
\au{Zichang Tan$^{3}$},
\au{Qi Yuan$^{6}$},
\au{Ruikun Yang$^{1}$},
\au{Benjia Zhou$^{1}$},
\au{Guodong Guo$^{7}$},
\au{Stan Z. Li$^{1,3,8}$}}

\address{
\add{1}{Faculty of Information Technology, M.U.S.T, Avenida WaiLong, Taipa, Macau, China}
\add{2}{School of computer and information technology, Beijing Jiaotong University, Beijing, China}
\add{3}{National Laboratory of Pattern Recognition, Institute of Automation, Chinese Academy of Science,  Beijing, China}
\add{4}{Universitat de Barcelona and Computer Vision Center, Barcelona, Spain}
\add{5}{Instituto Nacional de Astrof\'isica, \'Optica y Electr\'onica, Puebla,  Mexico and  Computer Science Department at  CINVESTAV-Zacatenco, Mexico.}
\add{6}{School of software, Beihang University,Beijing, China}
\add{7}{Institute of Deep Learning, Baidu Research and National Engineering Laboratory for Deep Learning Technology and Application, Beijing, China}
\add{8}{Westlake University, Hangzhou, China}
\email{jun.wan@ia.ac.cn}
}


\begin{abstract}
Face anti-spoofing is critical to prevent face recognition systems from a security breach. The biometrics community has 
achieved impressive progress recently due  the excellent performance of deep neural networks and the availability of large datasets. Although ethnic bias has been verified to severely affect the performance of face recognition systems, it still remains an open research problem in face anti-spoofing. Recently, a multi-ethnic face anti-spoofing dataset, CASIA-SURF CeFA, has been released with the goal of measuring the ethnic bias. It is the largest up to date cross-ethnicity face anti-spoofing dataset covering $3$ ethnicities, $3$ modalities, $1,607$ subjects, 2D plus 3D attack types, and the first dataset including explicit ethnic labels among the recently released datasets for face anti-spoofing. We organized the Chalearn Face Anti-spoofing Attack Detection Challenge which consists of single-modal (\eg, RGB) and multi-modal (\eg, RGB, Depth, Infrared (IR)) tracks around this novel resource to boost research aiming to alleviate the ethnic bias. Both tracks have attracted $340$ teams in the development stage, and finally $11$ and $8$ teams have submitted their codes in the single-modal and multi-modal face anti-spoofing recognition challenges, respectively.  All the results were verified and re-ran by the organizing team, and the results were used for the final ranking. This paper presents an overview of the challenge, including its design, evaluation protocol and a summary of results. We analyze the top ranked solutions and draw conclusions derived from the competition. In addition we outline future work directions.
\end{abstract}
\maketitle

\section{Introduction}\label{sec1}
Face anti-spoofing aims to determine whether the captured face from a face recognition system is real or fake. It is essential to protect face recognition systems from malicious attacks, such as a printed face photograph (\ie, print attack), displaying videos on digital devices (\ie, replay attack), or even 3D attacks (\ie, face mask). Therefore, the presentation attack detection (PAD) task is a vital stage prior to visual face recognition systems which has been widely applied in financial payment, access control, phone unlocking and surveillance. Some early temporal-based face PAD works~\cite{Pan2007Eyeblink,wang2009face,kollreider2008verifying,Bharadwaj2013Computationally} attempt to detect the evidence of liveness (\eg, eye-blinking), which require a constrained human interaction. However, these methods become vulnerable if someone presents a replay attack or a print photo attack with cut eye/mouth regions. Another works are based on static texture analysis~\cite{Pan2011Monocular,Komulainen2014Context}. However, these algorithms are not accurate enough because of the use of handcrafted features, such as LBP~\cite{chingovska2012effectiveness,Yang2013Face,Maatta2012Face}, HoG~\cite{Yang2013Face,Maatta2012Face,schwartz2011face} and GLCM~\cite{schwartz2011face}, that not necessarily are able to characterize samples, and adopt traditional classifiers such as SVM and LDA, which may be limited given the complexity of the task. Recently, CNN-based face PAD methods~\cite{feng2016integration,li2016original,Patel2016Secure,yang2014learn, Liu2018Learning,Jourabloo2018Face} have shown impressive progress due to 
the excellent performance of deep neural networks~\cite{Krizhevsky2012ImageNet,yang2014learn,feng2016integration,Liu2018Learning} and the availability of large datasets~\cite{Chingovska2012On,Zhang2012A,Boulkenafet2017OULU,Liu2018Learning,DBLP:conf/cvpr/abs-1812-00408,li2020casiasurf}. Although these methods achieve near-perfect performance in intra-database experiments, they are still vulnerable when facing complex authentication scenarios. In particular, ethnic bias  has been verified to severely affect the performance of face recognition systems~\cite{race_bias_fr,AlviTurning,Wang_2019_ICCV}, representing an open research problem in face anti-spoofing.
\begin{figure}[t]
	\begin{center}
	\includegraphics[width=1.0\linewidth]{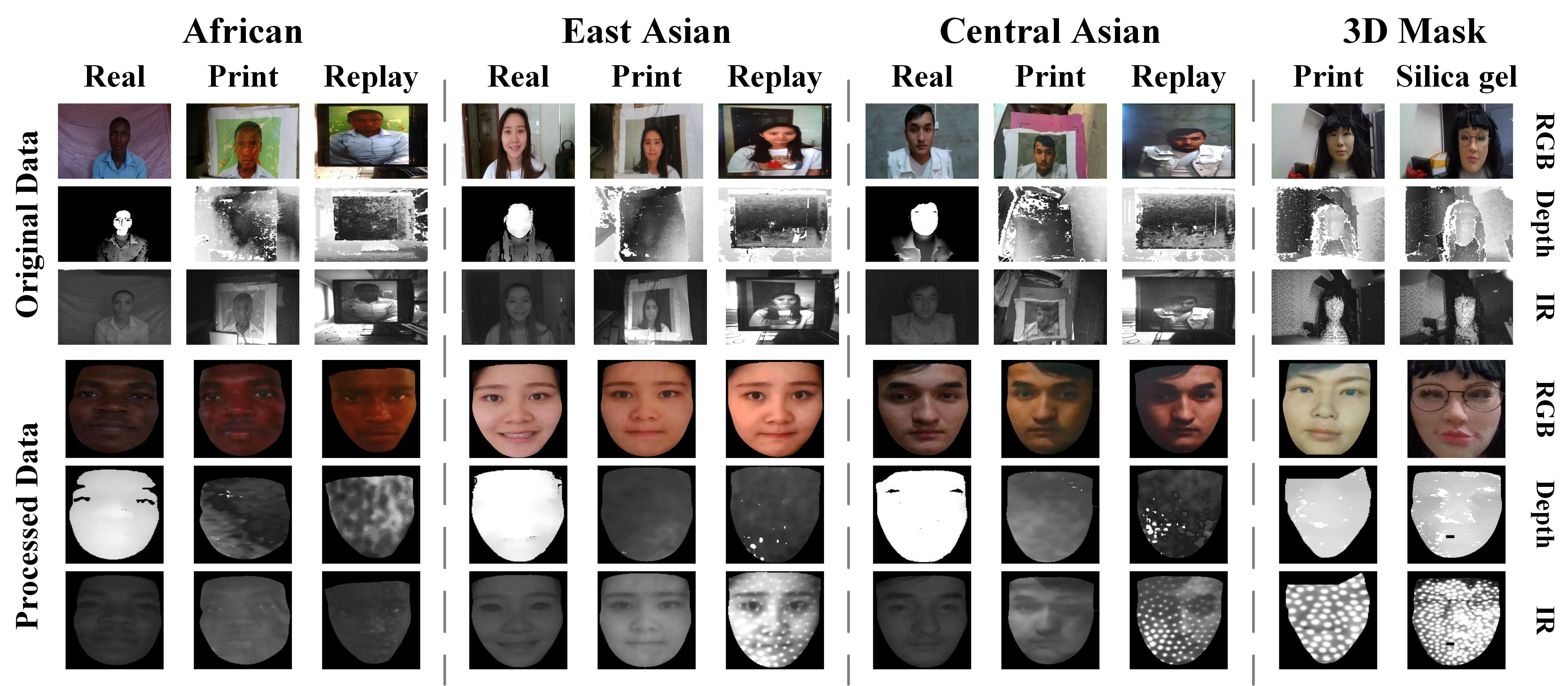}
	\end{center}
	\caption{Samples of the CASIA-SURF CeFA dataset. It contains $1,607$ subjects, $3$ different ethnicities (\ie, Africa, East Asia, and Central Asia), with $4$ attack types (\ie, print attack, replay attack, 3D print and silica gel attacks).}
	\label{samp_data}
\end{figure}

\begin{table}[!b]
\processtable{Team and affiliations name are listed in the final ranking of this challenge(Single-modal)\label{affiliations_Single}}
{\begin{tabular*}{20pc}{@{\extracolsep{\fill}}lll@{}}\toprule Ranking
& Team Name                  & Leader Name, Affiliation \\ \midrule
1  & VisionLabs              & Alexander Parkin, visionlabs \\
2  & BOBO                    & Zitong Yu, OULU unv.\\
3  & Harvest                 & Jiachen Xue, Horizon \\
4  & ZhangTT                 & Zhang Tengteng, CMB \\
5  & Newland\_tianyan        & Xinying Wang, Newland Inc. \\
6  & Dopamine                & Wenwei Zhang, huya \\
7  & IecLab                  & Jin Yang, HUST \\
8  & Chuanghwa Telecom Lab.  & Li-Ren Hou, Chunghwa Telecom \\
9  & Wgqtmac                 & Guoqing Wang, ICT \\
10  & Hulking                & Yang, Qing, Intel \\
11  & Dqiu                   & Qiudi \\
\botrule
\end{tabular*}}{}
\end{table}

In fact, we have verified in our previous work~\cite{li2020casiasurf} that state-of-the-art (SOTA) PAD algorithms do suffer from severe ethnic bias, for example, 
the ACER metric values vary widely on the test samples with different ethnicities for the same algorithm. In order to alleviate the ethnic bias and ensure that face PAD methods are in a safe reliable condition for users of different ethnicities, Liu~\etal~\cite{li2020casiasurf} introduce the largest up to date  Cross-ethnicity Face Anti-spoofing (CeFA) dataset, covering $3$ ethnicities, $3$ modalities, $1,607$ subjects, and 2D plus 3D attack types.  Some samples of the CASIA-SURF CeFA dataset are shown in Fig.~\ref{samp_data}. Four protocols are defined to measure the effect under varied evaluation conditions, such as cross-ethnicity, unknown spoofs or both of them. To the best of our knowledge, CeFA is the first dataset including explicit ethnic labels in current published datasets for face anti-spoofing. Additionally, they provide a baseline which includes two parts to alleviate above bias: (1) The static-dynamic fusion mechanism applied in each modality (\ie, RGB, Depth and infrared image); (2) The partially shared fusion strategy is proposed to learn complementary information from multiple modalities.

Leveraging on the CeFA dataset, we organized the \emph{Chalearn Face Anti-spoofing Attack Detection Challenge} comprising \emph{single-modal (\eg, RGB) and multi-modal (\eg, RGB, Depth, IR) tracks} collocated with the Workshop on Media Forensics at CVPR2020. The goal of this challenge was to boost research  on facial PAD aiming to alleviate the ethnic bias. Both tracks,  single-modal track\footnote{\url{https://competitions.codalab.org/competitions/22151}} and multi-modal track~\footnote{\url{https://competitions.codalab.org/competitions/22036}}  were run simultaneously on the Codalab platform.  The competition attracted  $340$ teams in the development stage, with  $11$ and $8$ teams entering the final evaluation stage for the single-modal and multi-modal face anti-spoofing recognition tracks, respectively. A summary with the names and affiliations of teams that entered the final stage is shown in Tables~\ref{affiliations_Single} and~\ref{affiliations_multi} for the single-modal  and multi-modal tracks, respectively.
\begin{table}[!b]
	\processtable{Team and affiliations name are listed in the final ranking of this challenge(Multi-modal)\label{affiliations_multi}} {\begin{tabular*}{20pc}{@{\extracolsep{\fill}}lll@{}}\toprule
		Ranking  & Team Name  & Leader Name, Affiliation \\
			\midrule
			1  & BOBO  & Zitong Yu, OULU unv. \\
			2  & Super  & Zhihua Huang, USTC\\
			3  & Hulking &  Qing Yang, Intel  \\
			4  & Newland\_tianyan  & Zebin Huang, Newland Inc. \\
			5  & ZhangTT  &Tengteng Zhang, CMB \\
			6  & Harvest  &Yuxi Feng, Horizon\\
			7  & Qyxqyx  &Yunxiao Qin, NWPU \\
			8  & Skjack  &Sun Ke, XMU \\
			\botrule
	\end{tabular*}}{}
\end{table}

Compared to previous challenges on related topics~\cite{chakka2011competition,chingovska20132nd,boulkenafet2017competition,ChallengeCVPR2019}, the algorithms of all participating teams were based on deep learning and did not require of external resources like datasets and pre-trained models. This was a rule established in the challenging that not only provides a fairer evaluation scenario, 
but also brings benefits for reproduciblity and algorithm implementation in practical applications. To sum up, the contributions of this paper are summarized as follows:
\begin{itemize}
    \item We describe the design and organization of both tracks of the \emph{Chalearn Face Anti-spoofing Attack Detection Challenge}, which is based on the CASIA-SURF CeFA dataset and was run on the CodaLab platform.
    \item  We provide a complete description of solutions developed in the context of the challenge.
    \item We point out critical points on face anti-spoofing detection  by comparing essential differences between a real face and a fake one from multiple aspects, also discussing future lines of research in the field.
\end{itemize}

\section{Challenge Overview}\label{sec2}
In this section, we describe the organized challenge, including a brief introduction to the CASIA-SURF CeFA dataset,  evaluation metrics, and the challenge protocol.

\subsection{CASIA-SURF CeFA}\label{sec2.1}
CASIA-SURF CeFA~\cite{li2020casiasurf} is the largest up to date cross-ethnicity face anti-spoofing dataset, covering $3$ ethnicities, $3$ modalities, $1,607$ subjects, and 2D plus 3D attack types. More importantly, it is the first public dataset designed for exploring the impact of cross-ethnicity in the study of face anti-spoofing. Some samples of the CASIA-SURF CeFA dataset are shown in Fig.~\ref{samp_data}.

The motivation of CASIA CeFA dataset is to provide a benchmark to allow for the evaluation of the generalization performance of new PAD methods. Concretely, four protocols are introduced to measure the affect under varied evaluation conditions: cross-ethnicity (Protocol 1), (2) cross-PAI (Protocol 2), (3) cross-modality (Protocol 3) and (4) cross-ethnicity and cross-PAI (Protocol 4). In order to facilitate the competition more challenging, we adopt Protocol $4$ in this challenge, which is a challenging protocol designed via combining conditions of both Protocol $1$ and $2$. As shown in Table~\ref{tab:PandS}, it has three data subsets: training, validation and testing sets, which contain $200$, $100$, and $200$ subjects for each ethnicity, respectively. Note that the remaining 107 subjects are 3D masks. 
In order to fully measure the cross-ethnicity performance of the algorithm, one ethnicity is used for training and validation, and the left two ethnicities are used for testing. Since there are three ethnicities in CASIA-SURF CeFA, a total of 3 sub-protocols (\ie, $4\_1$, $4\_2$ and $4\_3$ in Table~\ref{tab:PandS}) are adopted in this challenge. 
In addition to the ethnic variation, the factor of PAIs are also considered in this protocol by setting different attack types in training and testing phases.

\begin{table}[]
\processtable{Protocols and Statistics. Note that the A, C and E are short for Africa, Central Asia and East Asia, respectively. Track(S/M) means the Single/Multi-modal track. The PAIs means the presentation attack instruments.\label{tab:PandS}}
{
\scalebox{0.76}{
\begin{tabular}{@{}ccccccccccc@{}}
\toprule
\multicolumn{2}{c}{Track} &\multirow{2}{*}{Subset} &\multirow{2}{*}{Subjects(one ethnicity)} & \multicolumn{3}{c}{Ethnicity} &\multirow{2}{*}{PAIs} &\multicolumn{3}{c}{\# Num.img(rgb)} \\
\cmidrule(r){1-2}
\cmidrule(lr){5-7}
\cmidrule(l){9-11}
S & M &
& & 4\_1 & 4\_2  & 4\_3
& & 4\_1 & 4\_2  & 4\_3
\\ \midrule
\multirow{3}{*}{} & \multirow{3}{*}{\begin{tabular}[c]{@{}c@{}} \end{tabular}} & Train
& 1-200  & A   & C  & E  & Replay
& 33,713  & 34,367      & 33,152     \\
& & Valid & 201-300  & A   & C   & E  & Replay  & 17,008 & 17,693 & 17,109     \\
& & Test  & 301-500  & C\&E & A\&E & A\&C & Print  & 105,457  & 102,207 & 103,420 \\
\bottomrule
\end{tabular}
}
}{}
\end{table}

\subsection{Evaluation metric}
In this challenge we selected the recently standardized ISO/IEC 30107-3\footnote{\url{https://www.iso.org/obp/ui/iso}}  metrics: Attack Presentation Classification Error Rate (APCER), Normal Presentation Classification Error Rate (NPCER) and Average Classification Error Rate (ACER) as the evaluation metrics, these are defined as follows:
\begin{equation}APCER=FP/\left(FP+TN\right )\end{equation}
\begin{equation}NPCER=FN/\left (FN+TP\right )\end{equation}
\begin{equation}ACER=\left ( APCER+NPCER \right )/2\end{equation}
where TP, FP, TN and FN correspond to true positive, false positive, true negative and false negative, respectively. APCER and BPCER are used to measure the error rate of fake or live samples, respectively. Inspired by face recognition, the Receiver Operating Characteristic (ROC) curve is introduced for large-scale face Anti-spoofing detection in CASIA-SURF CeFA dataset, which can be used to select a suitable threshold to trade off the false positive rate (FPR) and true positive rate (TPR) according to the requirements of real applications. 

\subsection{Challenge protocol}
The challenge was run in the CodaLab platform, and comprised two stages as follows:
\subsubsection{Development Phase}
~(\emph{Started: Dec. 13, 2019 - Ended: in March 1, 2020}). During this phase, participants had access to labeled training subset and unlabeled validation subset. Since the protocol used in this competition (Protocol 4) comprises $3$ sub-protocols (see~\ref{sec2.1}), participants firstly need to train a model for each sub-protocol, then predict the score of the corresponding validation set, and finally simply merge the predicted scores and submit them to the CodaLab platform and receive immediate feedback via a public leader board.

\subsubsection{Final phase}
~(\emph{Started: March 1, 2020 - Ended: March 10, 2020}). During this phase, labels for the validation subset and the unlabeled testing subset were released. Participants can firstly take the labels of the validation subset to select a model with better performance, then they use this model to predict the scores of the corresponding testing subset samples, and finally submit the score files in the same way as the development phase. We will feedback all results of the three sub-protocols online which including the values of APCER, BPCER, and ACER. Like~\cite{Boulkenafet2017OULU}, the mean and variance of evaluated metrics for these three sub-protocols are calculated for final results. Note that to fairly compare the performance of participants' algorithms, this competition does not allow the use of other training datasets and pre-trained models. To be eligible for prizes, winners had to publicly release their code under a license of their choice and provide a fact sheet describing their solution.

\section{Description of solutions}\label{sec3}
In the final ranking stage, there were 19 teams submitting their code and face sheets for evaluation. According to the information provided, in the following we describe the solutions developed by each of the  teams, with  detailed descriptions 
for top ranked participants in both single-modal (RGB) and multi-modal (RGB-Depth-IR) face anti-spoofing recognition challenge tracks.

Tables~\ref{affiliations_Single} and~\ref{affiliations_multi} show the final ranking for both tracks. It can be seen from these tables that  most  participants came from the industrial community. Interestingly, the VisionLabs team was not only the winner of the single-modal track, but also the winner of the \emph{Chalearn LAP multi-modal face anti-spoofing attack detection challenge at CVPR 2019}~\cite{ChallengeCVPR2019}. In addition, the BOBO team designs novel central difference convolution (CDC)~\cite{yu2020searching} and Contrastive depth loss (CDL)~\cite{wang2020deep} for feature learning, and achieved second and first place in both single-modal and multi-modal tracks, respectively.  

\subsection{Single-modal Face Anti-spoofing Challenge Track}\label{sec3.1}
\begin{figure}[t]
	\begin{center}
	\includegraphics[width=1.0\linewidth]{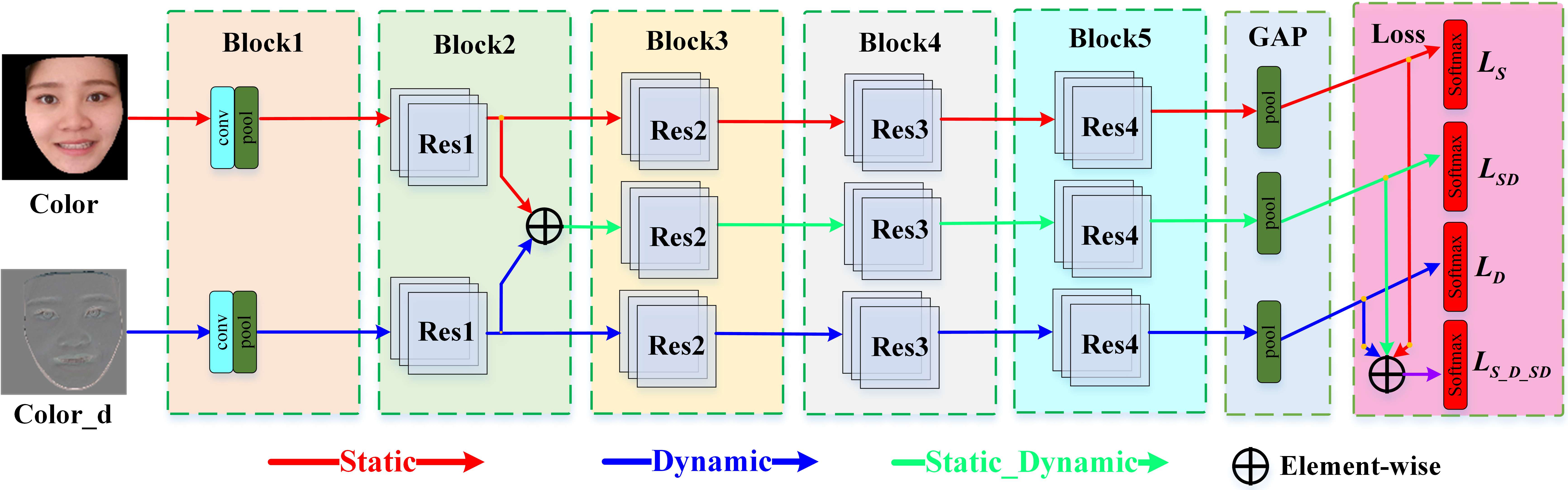}
	\end{center}
	\caption{The framework of SD-Net. The figure is provided by the baseline team and ranked $NO.11$ in single-modal track.}
	\label{fig:SD-Net}
\end{figure}

\noindent \textbf{Baseline.}~We provided a baseline for approaching this task via designing a SD-Net~\cite{li2020casiasurf} which takes Resnet18~\cite{he2016deep} as the backbone. As shown in Fig.~\ref{fig:SD-Net}, it contains $3$ branches: static, dynamic and static-dynamic branches, which learn hybrid features from static and dynamic images. For static and dynamic branches, each of them consists of $5$ blocks (\ie, conv, res1, res2, res3, res4) and $1$ Global Average Pooling (GAP) layer, while in the static-dynamic branch, the conv and res1 blocks are removed because it takes fused features of res1 blocks from static and dynamic branches as input.

For dynamic image generation, a detailed description is provided in~\cite{li2020casiasurf}. In short, we compute its dynamic image online with rank pooling using $K$ consecutive frames. Our selection of dynamic images for rank pooling in SD-Net is further motivated by the fact that dynamic images have proved its superiority to regular optical flow~\cite{wang2017ordered,fernando2017rank}.

\begin{figure}[t]
	\begin{center}
	\includegraphics[width=1.0\linewidth]{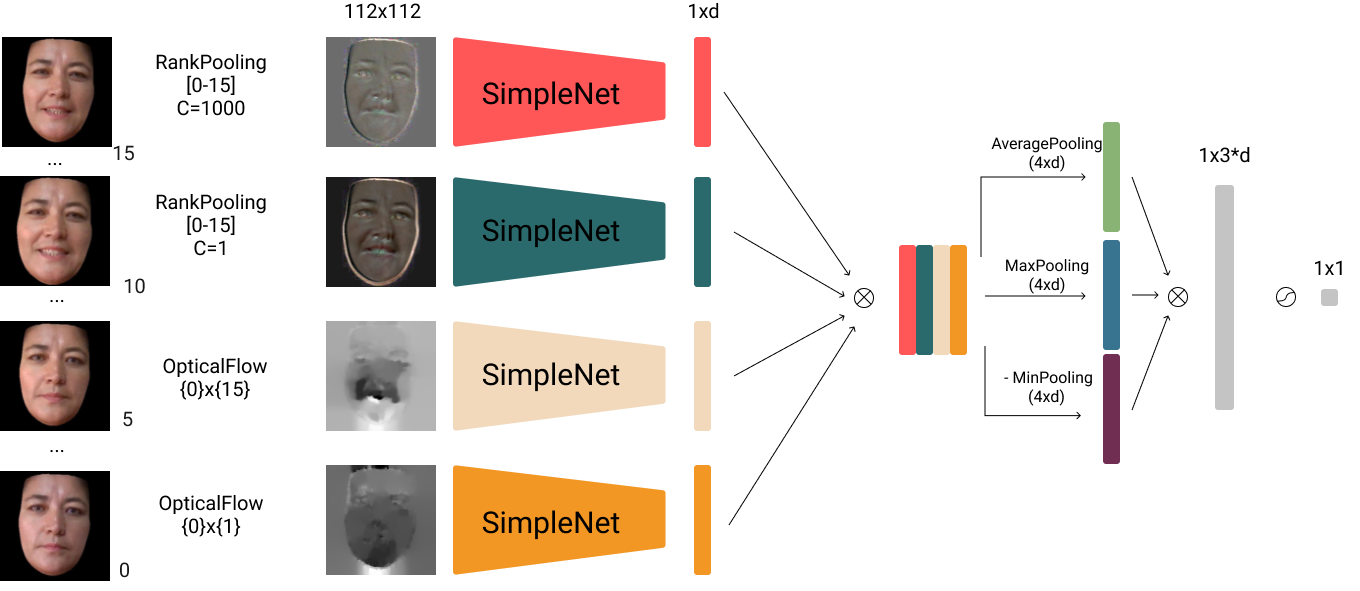}
	\end{center}
	\caption{The framework is provided by the VisionLabs team. Note that the SimpleNet architecture: 4 blocks of Conv $3\times3$- BatchNorm - Relu - MaxPool of sizes 16, 32, 64, 128 followed by Conv $5\times5$ with 256 filters. The figure is provided by the VisionLabs team and ranked $NO.1$ in single-modal track.}
	\label{fig:VisionLabs}
\end{figure}

\noindent \textbf{VisionLabs.}~Due to high differences in the train and test subsets (\ie, different ethnics and attack types), the VisionLabs team used a data augmentation strategy to help train robust models. Similar to previous works which convert RGB data to HSV and YCbCr color spaces~\cite{Boulkenafet2016Face}, or Fourier spectrum~\cite{Li2004Live}, they decided to convert RGB to other ``modalities'', which contain more authentic information instead of identity features. Specially, the Optical Flow and RankPooling are used as shown in Fig.~\ref{fig:VisionLabs}.

As shown in Fig.~\ref{fig:VisionLabs}, the proposed architecture  consists of four branches where two branches are used for dynamic images via dynamic pooling algorithm and the left two branches are used for the optical flow images. For optical flow modality, they calculated two flows between first and last images of RGB video as well as between the first and second images. For the rank pooling modality, they used the rank pooling algorithm~\cite{fernando2017rank} where different hyperparameters used to generate two different dynamic images.

Formally, a RGB video with $K$ frames is represented by $\{X_i^k\}$, where $i=0,...,K-1$ and $t=\{0,1\}$ is the label ($0$-fake, $1$-real). Then for each RGB video, they sample $L=16$ images uniformly, obtaining $\{X_j^k\}$, where $j=0, ..., 15$.
Then, they remove black borders and pad image to be square of size (112, 112). Then they apply intensive equal color jitter to all images, emulating different skin color.

As shown in Fig.~\ref{fig:VisionLabs}, they apply 4 ``modality'' transforms: rank pooling ($\{X_j^k\}$, C=1000), rank pooling ($\{X_j^k\}$, C=1), Flow($X_0^k$, $X_{15}^k$), Flow ($X_0^k$, $X_1^k$), where C is the hyperparameter for SVM in the rank pooling algorithm~\cite{fernando2017rank}. The code of rank pooling was released~\footnote{\url{https://github.com/MRzzm/rank-pooling-python}}. These transforms return 4 tensors with sizes $3\times112\times112$, $3\times112\times112$, $2\times112\times112$, $2\times112\times112$ respectively. Further, the features of each modal sample are extracted by an independent network (namely SimpleNet and its structure depicted in Fig.~\ref{fig:VisionLabs}) with size of $d=256$ and all features are concatenated to get a tensor of shape $4\times d$. Then they apply Max, Avg and Min pooling among first dimension and concatenate results to get $3 \times d$ tensor. Finally, a binary cross-entropy is adopted in their network. The code of VisionLabs was released~\footnote{\url{https://github.com/AlexanderParkin/CASIA-SURF_CeFA}}.

\begin{figure*}[t]
	\begin{center}
	\includegraphics[width=1.0\linewidth]{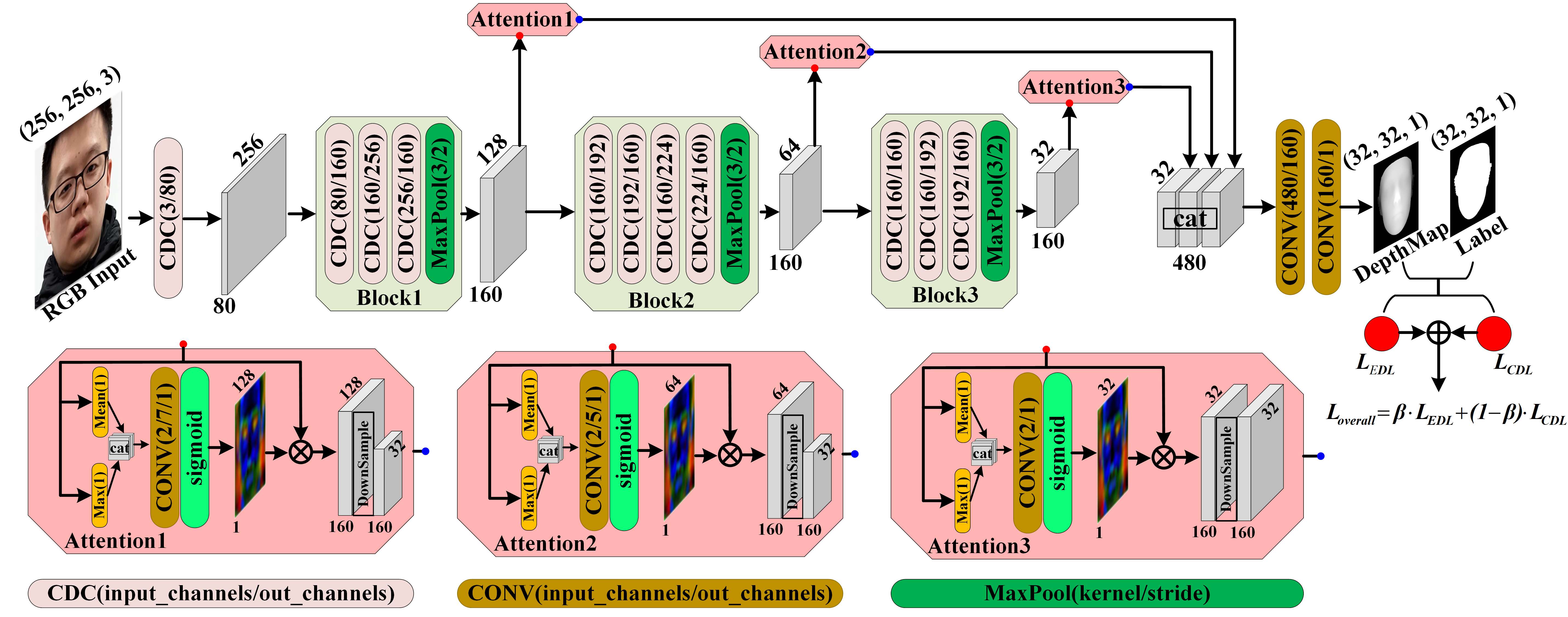}
	\end{center}
	\caption{The framework of regression network. The figure is provided by the BOBO team and ranked $NO.2$ in single-modal track.}
	\label{fig:BOBO}
\end{figure*}

\noindent \textbf{BOBO.}~Most CNN-based methods~\cite{patel2016cross,feng2016integration,li2016original,yang2014learn} only treat face anti-spoofing as a binary classification, and train the neural network supervised by the softmax loss. However, these methods fail to explore the nature of spoof patterns~\cite{Liu2018Learning}, which consist of skin detail loss, color distortion, moire pattern, motion pattern, shape deformation and spoofing artifacts. In order to relieve the above issues, similar to~\cite{wang2020deep}, the BOBO team adopts depth supervision instead of binary softmax loss for face anti-spoofing. Different from~\cite{wang2020deep}, they design a novel Central Difference Convolution (CDC)~\cite{yu2020searching} and a Contrastive Depth Loss (CDL) for feature learning and representation.

The structure of the depth map regression network based on CDC is shown in Fig.~\ref{fig:BOBO}. It consists of $3$ blocks, $3$ attention layers connected after each block, and $3$ down-sampling layers followed by each attention layer. Inspired by the residual network, they use a short-cut connection, which is concatenating the responses of Low-level Cell (Block1), Mid-level Cell (Block2) and High-level Cell (Block3), and sending them to two cascaded convolutional layers for depth estimation. All convolutional layers use the CDC network which is followed by a batch normalization layer and a rectified linear unit (ReLU) activation function. The size of input image and regression depth map are $3\times256\times256$ and $1\times32\times32$, respectively. Euclidean Distance Loss (EDL) is used for pixel-wise supervision in this work which is formulated:
\begin{equation}
\begin{split}
{L}_{EDL} = & ||\textbf{\rm{D}}_{P} - \textbf{\rm{D}}_{G}||_{2}^{2},
\label{eq:euclidean_distance_loss}
\end{split}
\end{equation}
where $\textbf{\rm{D}}_{P}$ and $\textbf{\rm{D}}_{G}$ are the predicted depth and groundtruth depth, respectively.

\begin{figure}[t]
	\begin{center}
	\includegraphics[width=1.0\linewidth]{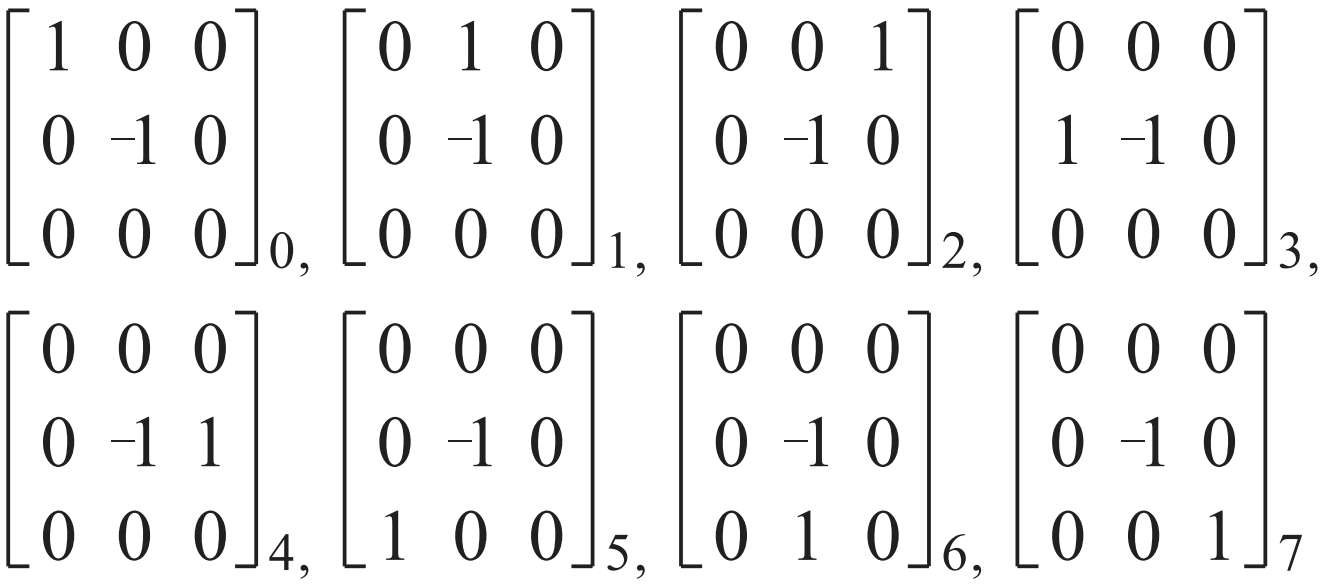}
	\end{center}
	\caption{The kernel $\rm{K}_{i}^{contrast}$ in contrastive depth loss.}
	\label{fig:kernal}
\end{figure}

EDL applies supervision on the predicted depth based on pixel one by one, ignoring the depth difference among adjacent pixels. Intuitively, EDL merely assists the network to learn the absolute distance between the objects to the camera. However, the distance relationship of different objects is also important to be supervised for the depth learning. Therefore, one proposed the Contrastive Depth Loss (CDL) to offer an extra supervision, which improves the generality of the depth-based face anti-spoofing model:
\begin{equation}
\small
\begin{split}
{{L}}_{CDL} = & \sum_i{||\textbf{\rm{K}}_{i}^{CDL} \odot \textbf{\rm{D}}_{P} - \textbf{\rm{K}}_{i}^{CDL} \odot \textbf{\rm{D}}_{G}||_{2}^{2}},
\label{eq:contrast_depth_loss}
\end{split}
\end{equation}
where $\textbf{\rm{K}}_{i}^{CDL}$ is the $i^{th}$ contrastive convolution kernel, $i \in [0,7]$. The details of the kernels can be found in Fig.~\ref{fig:kernal}.

Therefore, the total loss ${{L}_{overall}}$ employed by this team is defined as follow:
\begin{equation}
{{{L}}_{overall}} = \beta \cdot {{{L}}_{EDL}} + (1-\beta) \cdot {{L}}_{CDL},
\label{eq:mutil_loss}
\end{equation}
where $\beta$ is the hyper-parameter to trade-off EDL loss and CDL loss in the final overall loss ${{L}_{overall}}$. Finally their code is publicly available~\footnote{\url{https://github.com/ZitongYu/CDCN/tree/master/FAS_challenge_CVPRW2020}}.

\begin{figure}[t]
\begin{center}
\includegraphics[width=1.0\linewidth]{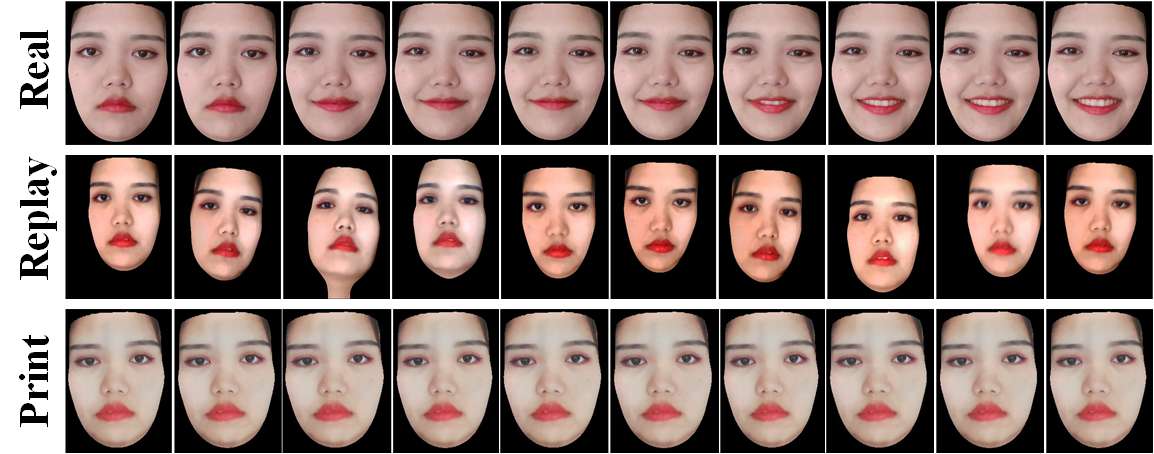}
\end{center}
\caption{Visual comparison of real face, replay attack, print attack motion information for Harvest team.}
\label{fig:harvest_1}
\end{figure}

\begin{figure}[t]
\begin{center}
\includegraphics[width=1.0\linewidth]{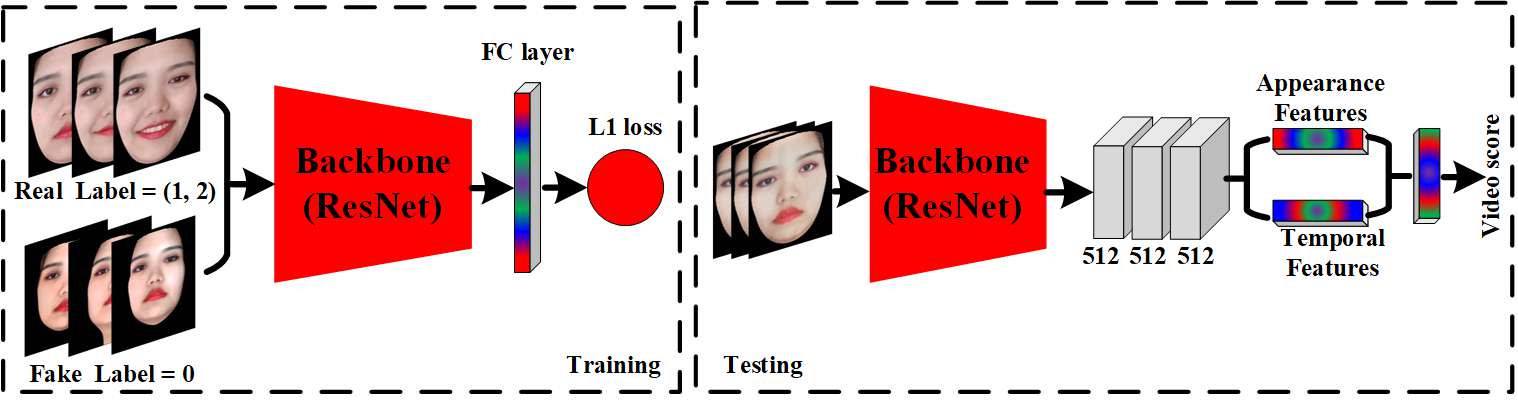}
\end{center}
\caption{The framework of training and testing phases for Harvest. The figure is provided by the Harvest team and ranked $NO.3$ in single-modal track.}
\label{fig:harvest_2}
\end{figure}

\noindent \textbf{Harvest.}~It can be observed from Table~\ref{tab:PandS} that the attack types of the spoofs in the training and testing subsets are different. The Harvest team considered the motion information of real faces is also an important discriminative cue for face anti-spoofing attack detection. Therefore, how to effectively learn the motion information of real faces from the interference motion information of the replay attack is a key step. As shown in Fig.~\ref{fig:harvest_1}, the live frame displays obvious temporal variations, especially in expressions, while there is very little facial changes in the print spoof samples for same subject, which inspires the Harvest team to capture the subtle dynamic variations by relabelling live sequence. Suppose the labels of spoof and live samples are 0 and 1 respectively. They define a new temporal-aware label via forcing the labels of the real face images in a sequence to change uniformly from 1 to 2, while the spoofing faces stay 0. Let $X=\{x_{1}, x_{2}, \ldots, x_{n}\}$ denote a video containing $n$ frames, where $x_{1}$ and $x_{n}$ represent the first and final frames, respectively. They encode this implicit temporal information by reformulating the ground-truth label, such as:
\begin{equation}
\begin{aligned}
\label{eq:P}
gt_{i}=1+\frac{i}{n}
\end{aligned}
\end{equation}
where the genuine label grows over time. Note that they do not encode the temporal variations in the spoof video due to their irregular variations in sequence. As shown in Fig.~\ref{fig:harvest_2}, the overall framework consists of two parts below:

(1) In the training stage, they encode inherent discriminative information by relabelling live sequence.

(2) In inference stage, they aggregate the static-spatial features with dynamic-temporal information for sample classification. Finally, combined with the strong learning ability of backbone, their method achieved $3rd$ in the single-modal track and the code is publicly available~\footnote{\url{https://github.com/yueyechen/cvpr20}}.

\begin{figure}[t]
\begin{center}
\includegraphics[width=1.0\linewidth]{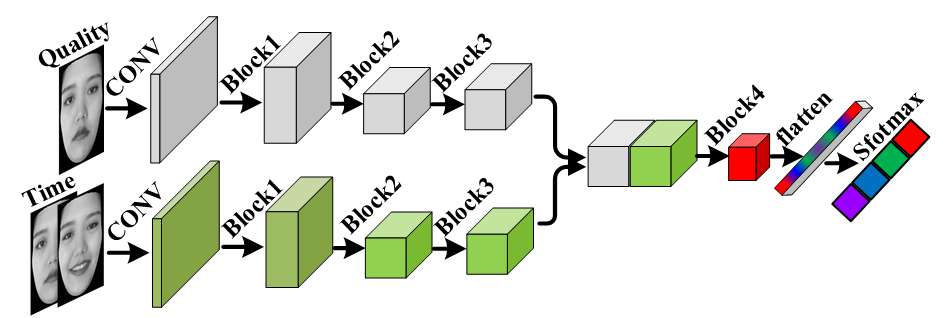}
\end{center}
\caption{Architecture of the proposed for single-modal track. The figure is provided by the ZhangTT team and ranked $NO.4$ in single-modal track.}
\label{fig:zhangTT1}
\end{figure}

\noindent \textbf{ZhangTT.}~Similar to the SD-Net in baseline~\cite{li2020casiasurf}, this team proposes a two-branch network to learn hybrid features from static and temporal images. They call it quality and time tensor, respectively. As shown in Fig.~\ref{fig:zhangTT1}, they take the ResNet~\cite{he2016deep} as backbone for each branch and use the single frame and multi-frame as input of the two branches. Specially, the quality tensor and time tensor are first sent to a normal $7\times7$ receptive field convolution layer for preliminary feature extraction. After feature extraction by three independent blocks, a higher-level expression quality feature map and time feature map were obtained. Then the quality feature and the time feature are concatenated together to form a new feature map for final classification with a binary cross-entropy loss function. The blocks in this work are same with the ResNet block~\cite{he2016deep}.

For data preprocessing, they first discarded the color information by converting the RGB modality to grayscale space, and then used histogram equalization to mitigate the skin tone gap between ethnicities. Finally, they adopted the following four strategies to reduce the difference between replay and print attacks: 1) They regard face anti-spoofing work as a classification task for 4 classes instead of binary. Such as the 4 categories are live-invariable (label 0), fake-invariable (label 1), live-variable (label 2), fake-variable (label 3), respectively. 2) Dithering each channel of the attack sample solves the problem of consistency of each frame of the print attack. 3) To enhance the robustness, consider randomly superimposing Gaussian noise and superimposing gamma correction on each channel of the time tensor. 4) In order to discriminate the texture difference, the first channel of the time tensor is separately identified and recorded as the quality tensor. It is sent to the network to extract features without noise superposition. Their code is publicly available~\footnote{\url{https://github.com/ZhangTT-race/CVPR2020-SingleModal}}.

\noindent \textbf{Newland\_tianyan.}~This team mainly explores single-modal track from two aspects of data augment and network design. For data augment, on the one hand, they introduced print attacks in the training set by randomly pasting paper textures on real face samples. On the other hand, they performed random rotation, movement, brightness transformation, noise and fold texture addition on the same frame of real face to simulate the case that there is no micro expression change for the print attack. For network design, this team uses a 5-layer sequence network which taking 16 frames of samples as input to learn the temporal features. In order to improve the generalization faced with different ethnicities, the images are subtracted from the neighborhood mean before sending to the network due to the samples of different ethnicities vary widely in skin color. Their code is publicly available~\footnote{\url{https://github.com/XinyingWang55/RGB-Face-antispoofing-Recognition}}.

\begin{figure}[t]
\begin{center}
\includegraphics[width=1.0\linewidth]{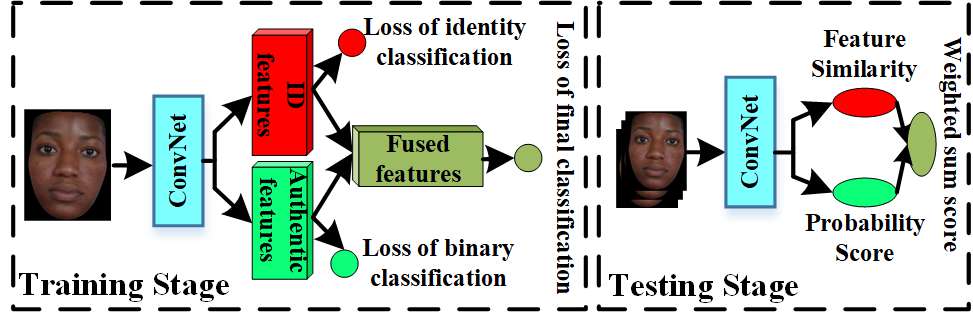}
\end{center}
\caption{The architecture of multi-task network for face anti-spoofing. The figure is provided by the Dopamine team and ranked $NO.6$ in single-modal track.}
\label{fig:Dopamine}
\end{figure}

\noindent \textbf{Dopamine.}~This team uses face ID information for face anti-spoofing tasks. The architecture is shown in Fig.\ref{fig:Dopamine}, a multi-task network is designed to learn the features of identity and authenticity simultaneously. In the testing phase, these two scores are combined to determine whether a sample is a real face. They use the softmax score from real/fake classifier and the feature computed by the backbone network (Resnet100) to compute minimal similarity between same person. In theory, the feature similarity score of the attack sample is close to 1, and the real face is close to 0. Their code is publicly available~\footnote{\url{https://github.com/xinedison/huya_face}}.

\noindent \textbf{IecLab.}~This team uses feathernet and 3DResNet~\cite{HaraCan} to learn the authenticity and expression features of the samples, and finally merged the two features for anti-spoofing tasks. Their code is publicly available~\footnote{\url{https://github.com/1relia/CVPR2020-FaceAntiSpoofing}}.

\noindent \textbf{Chuanghwa Telecom Lab.}~This team combines subsequence features with Bag of local features~\cite{shen2019facebagnet} within the framework of MIMAMO-Net\footnote{\url{https://github.com/wtomin/MIMAMO-Net}}. Finally, the ensemble learning strategy is used for feature fusion. Their code is publicly available~\footnote{\url{https://drive.google.com/open?id=1ouL1X69KlQEUl72iKHl0-_UvztlW8f_l}}.

\noindent \textbf{Wgqtmac.}~This team focused on improving face anti-spoofing generalization ability and proposed an end-to-end trainable face anti-spoofing approach based on deep neural network. They choose Resnet18~\cite{he2016deep} as the backbone and use warmup strategy to update the learning rate. The learned model performs well on the developing subset. However, it is easily overfitted on the training set and gets worse results on the testing set. Their code is publicly available~\footnote{\url{https://github.com/wgqtmac/cvprw2020.git}}.

\noindent \textbf{Hulking.}~The main role of PipeNet proposed by this team is to selectively and adaptively fuse different modalities for face anti-spoofing task. Since single-modal track only allow the use of RGB data, the team's method has limited performance in this challenge. We detail the team's algorithm in Section~\ref{Multi-modal-track}. Their code is publicly available~\footnote{\url{https://github.com/muyiguangda/cvprw-face-project}}.

\noindent \textbf{Dqiu.}~This team treats the face anti-spoofing as a binary classification task and uses Resnet50~\cite{he2016deep} as the backbone to learn the features. Since no additional effective strategies were used, no good results were achieved on the testing set.

 \begin{figure}[t]
	\begin{center}
	\includegraphics[width=1.0\linewidth]{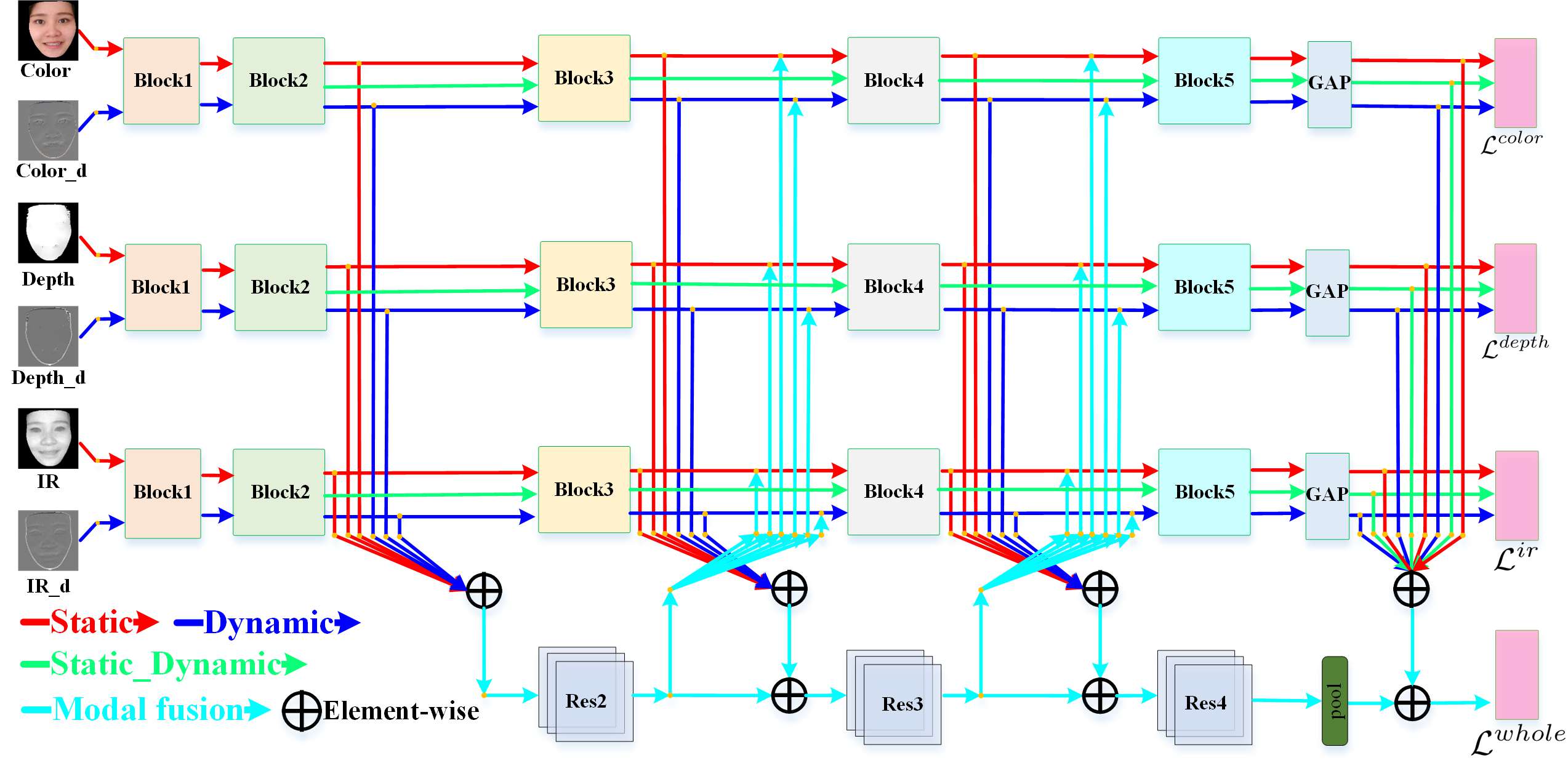}
	\end{center}
	\caption{The framework of PSMM-Net. The figure is provided by the baseline team and ranked $NO.8$ in multi-modal track.}
	\label{fig:PSMM-Net}
\end{figure}

\subsection{Multi-modal}\label{Multi-modal-track}
\noindent \textbf{Baseline.}~In order to take full advantage of multi-modal samples to alleviate the ethnic and attack bias, we propose a novel multi-modal fusion network, namely PSMM-Net~\cite{li2020casiasurf}. As shown in Fig.~\ref{fig:PSMM-Net}. It consists of two main parts: a) the modality-specific network, which contains three SD-Nets to learn features from RGB, Depth, IR modalities, respectively; b) and a shared branch for all modalities, which aims to learn the complementary features among different modalities. In order to capture correlations and complementary semantics among different modalities, information exchange and interaction among SD-Nets and the shared branch are designed.

There are two main kind of losses employed to guide the training of PSMM-Net. The first corresponds to the losses of the three SD-Nets, \ie color, depth and ir modalities, denoted as $\mathcal L^{color}$, $\mathcal L^{depth}$ and $\mathcal L^{ir}$, respectively. The second corresponds to the loss that guides the entire network training, denoted as $\mathcal L^{whole}$, which bases on the summed features from all SD-Nets and the shared branch. The overall loss $\mathcal L$ of PSMM-Net is denoted as:
\begin{equation}\label{Eq:multi_modality_loss}
\begin{split}
\mathcal{L}= \mathcal{L}^{whole} + \mathcal{L}^{color} +  \mathcal{L}^{depth} + \mathcal{L}^{ir}
\end{split}
\end{equation}

\begin{figure*}[t]
	\begin{center}
	\includegraphics[width=1.0\linewidth]{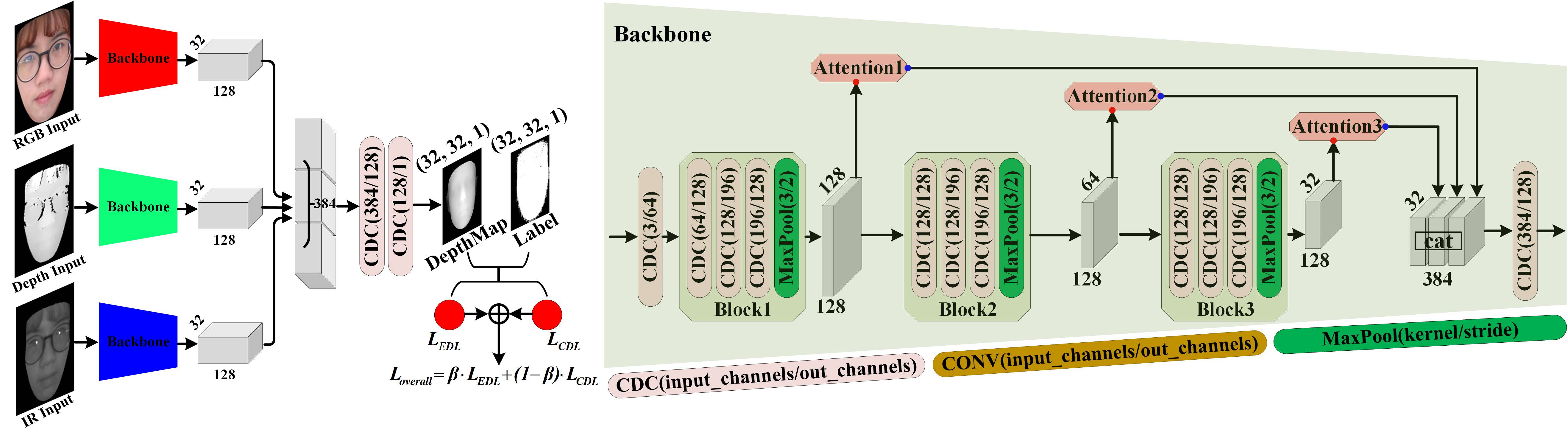}
	\end{center}
	\caption{The framework of regression network for $3$ modalities. The figure is provided by the BOBO team and ranked $NO.2$ in multi-modal track.}
	\label{fig:BOBO_2}
\end{figure*}

\noindent \textbf{BOBO.}~For the Multi-modal track, as shown in Fig.~\ref{fig:BOBO_2}, this team takes $3$ independent networks (Backbone) to learn the features of the $3$ modalities (\eg, RGB, Depth, IR). Therefore, the entire structure consists of two main parts: a) the modality-specific network, which contains three branches (the backbone network of each modality branch is not shared) to regress depth maps of RGB, Depth, IR modalities, respectively; b) a fused branch (via concatenation) for all modalities, which aims to learn the complementary features among different modalities and output final depth map with same size ($1\times32\times32$) of single-modal track. Similar to single-modal track, the CDL and CDE loss functions are used in multi-modal track in the form of weighted sums.
As the feature-level fusion strategy (see Fig.~\ref{fig:BOBO_2}) might not be optimal for all protocols, they also try two other fusion strategies: 1) input-level fusion via concatenating three-modal inputs to $256\times256\times9$ directly, and 2) score-level fusion via weighting the predicted score from each modality. For these two fusion strategies, the architecture of single-modal CDCN (see Fig.~\ref{fig:BOBO}) is used. Through comparative experiments, they concluded that the input-level fusion (\ie, simple fusion with concatenation) might be sub-optimal because it is weak in representing and selecting the importance of modalities. Therefore, this final result is combined with the best sub-protocols results (\ie, feature-level fusion for protocol 4\_1 while score-level fusion for protocol 4\_2 and 4\_3). Especially for score-fusion, they weight the results of RGB and Depth modalities averagely as the final score (\ie, $fusion\_score = 0.5 \times RGB\_score + 0.5 \times depth\_score $). This simple ensemble strategy helps to boost the performance significantly in their experiments.

\noindent \textbf{Super.}~CASIA-SURF CeFA is characterized by multi-modality (\ie, RGB, Depth, IR) and a key issue is how to fuse the complementary information between the three modalities. This team explored multi-modal track from three aspects: (1) Data preprocessing. (2) Network construction. (3) Ensemble strategy design.

Since the dataset used in this competition retained the black background area outside the face, this team tried to remove the background area using the histogram threshold method to mitigate its interference effect on model learning. To increase the diversity of training samples, they use random rotation within the range of [$-30^0$, $30^0$], flipping, cropping and color distortion for data augmentation. Note that the three modalities of the same sample are maintained in a consistent manner to obtain the features of the corresponding face region.

\begin{figure}[t]
\begin{center}
\includegraphics[width=1.0\linewidth]{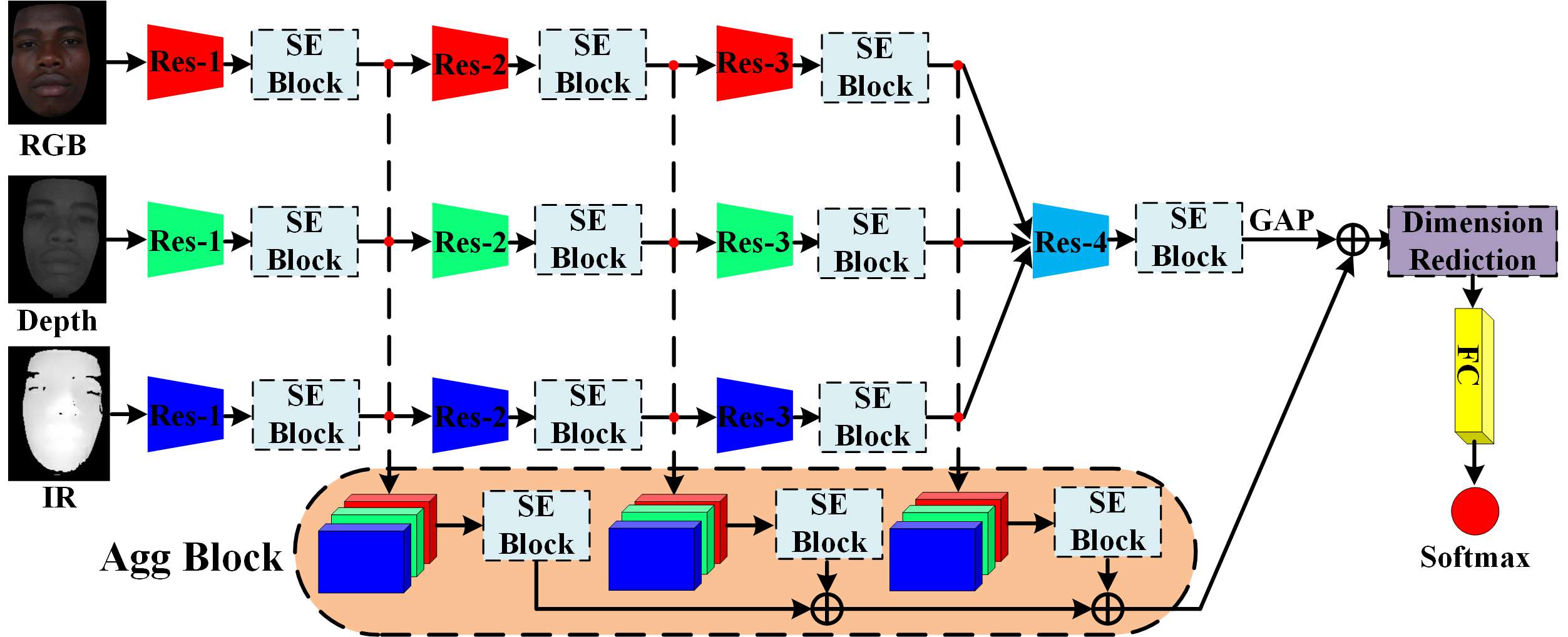}
\end{center}
\caption{The framework of Super team. The ResNet34 or IR\_ResNet50 as the backbone. The figure is provided by the Super team and ranked $NO.2$ in multi-modal track.}
	\label{fig:super_net}
\end{figure}

Inspired by~\cite{DBLP:conf/cvpr/abs-1812-00408} which employs the ``Squeeze-and-Excitation'' Block (SE Block)~\cite{hu2018senet} to re-weighting the hierarchy features of each modality, this team takes a multi-stream architecture with three subnetworks to study the dataset modalities, as shown in Fig.\ref{fig:super_net}. We can see that the RGB, Depth and IR data are learnt separately by each stream, and then shared layers are appended at a point (Res-4) to learn joint representations. However, the single-scale SE block~\cite{hu2018senet} does not make full use of features from different levels. To this end, they extend the SE fusion from single scale to multiple scales. As shown in Fig.~\ref{fig:super_net}, the Res-1, Res-2 and Res-3 blocks from each stream extract features from different modalities. After that, they first fuse features from different modalities via the SE block after Res-1, Res-2 and Res-3 respectively, then concatenate these fused features and sending them to aggregation block (Agg Blcok), next merging these features (including shared branch features after the Global Average Pooling (GAP)) via element summation operations similar to~\cite{parkin2019recognizing}. Finally, they use the merged features to predict real and fake. Differently from~\cite{parkin2019recognizing}, they add a dimension reduction layer before fully-connected (FC) layer for avoiding the overfitting.

To increase the robustness to unknown attack types and ethnicities, they design several new networks based on the basic network shown in Table~\ref{tab:ensemble ways}. Such as the Network A with a dimension reduction layer and without SE fusion after each res block. While the Network B and C are similar to~\cite{parkin2019recognizing} and~\cite{DBLP:conf/cvpr/abs-1812-00408} respectively. For the IR\_ResNet50, it uses the improved residual block which aims at fitting the face recognition task. In the experiments, 
they found that different networks performed differently under the same sub-protocol. 
Therefore, they selectively trained these networks according to different sub-protocols and get the final score via averaging the results of selected networks. Their code is publicly available~\footnote{\url{https://github.com/hzh8311/challenge2020_face_anti_spoofing}}.
\begin{table}[]
\processtable{The networks ensemble ways adopted by Super team. Each network carries functions marked by \checkmark.
\label{tab:ensemble ways}}
{
\scalebox{1.0}{
\begin{tabular}{@{}ccccc@{}}
\toprule
Network & Backbone     & SE block & Dimension reduction & Agg block \\
\midrule
A       & ResNet34     &             & \checkmark      & \checkmark       \\
B       & ResNet34     & \checkmark  &                 & \checkmark       \\
C       & ResNet34     & \checkmark  &                 &                  \\
D       & IR\_ResNet50 &             &                 & \checkmark       \\
\bottomrule
\end{tabular}
}
}{}
\end{table}

\begin{figure}
	\begin{center}
		\centering
		\includegraphics[width=1.0\linewidth]{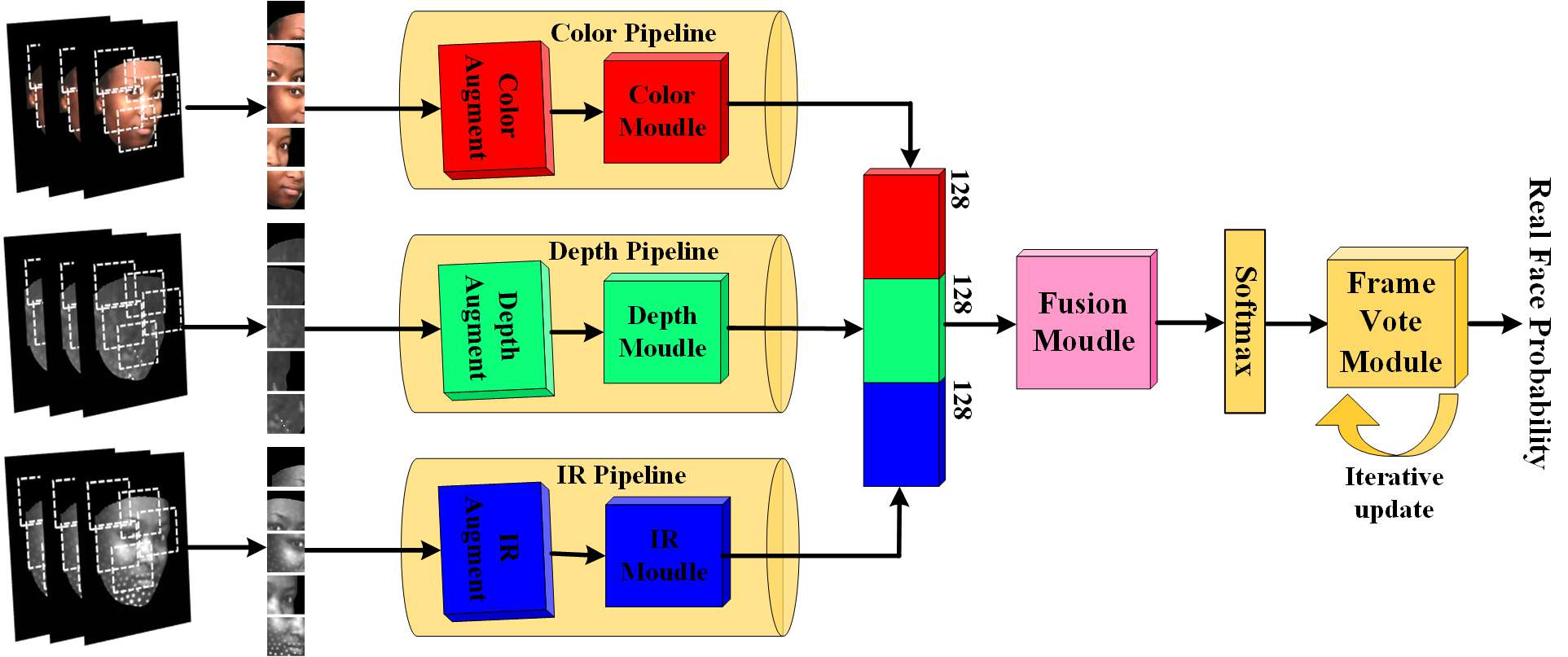}
	\end{center}
	\caption{The overall architecture of PipeNet. The figure is provided by the Hulking team and ranked $NO.3$ in multi-modal track.}
\label{fig:PipeNet}
\end{figure}

\noindent \textbf{Hulking.}~As for this team, they propose a novel Pipeline-based CNN (namely PipeNet) fusion architecture which taking modified SENet-154~\cite{hu2018senet} as backbone for multi-modal face anti-spoofing. Specifically, as shown in Fig.~\ref{fig:PipeNet}, it contains two modules, namely SMP (Selective Modal Pipeline) module and LFV (Limited Frame Vote) module for the input of multiple modalities and sequence video frames, respectively. We can see that the framework contains three SMP modules, and each module takes a modal data (\ie, RGB, Depth, IR) as input. Taking the RGB modality as an example, they firstly use one frame as input and randomly crop it into patches, then send them to $Color Pipeline$ which consists of data augmentation and feature extraction operations. They use a fusion strategy, which is concatenating the responses of $Color Pipeline$, $Depth Pipeline$ and $IR Pipeline$, and sending them to $Fusion Moudle$ for further feature abstraction. After the linear connection, input all frame features of the video to the LFV module, and iteratively calculate the probability that each frame sample belongs to the real face. Finally, the output is prediction for real face probability of the input face video.

\noindent \textbf{Newland\_tianyan.}~For multi-modal track, this team uses two independent ResNet-9~\cite{he2016deep} as backbones to learn the features of depth and ir modal data respectively. Similar to single-modal track, the inputs of depth branch are subtracted from the neighborhood mean before entering the network. In addition to data augment similar to the single-modal track, they transferred the RGB data of real samples to gray space and added light spots for data augment. Their code is publicly available~\footnote{\url{https://github.com/Huangzebin99/CVPR-2020}}.

\begin{figure}[t]
\begin{center}
\includegraphics[width=1.0\linewidth]{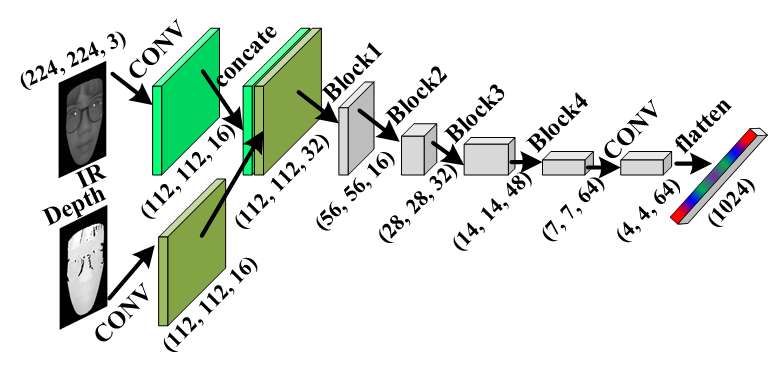}
\end{center}
\caption{Architecture of the proposed for multi-modal track. The figure is provided by the ZhangTT team and ranked $NO.5$ in multi-modal track.}
\label{fig:zhangTT2}
\end{figure}

\noindent \textbf{ZhangTT.}~A multi-stream CNN architecture called ID-Net is proposed for multi-modal track. Since the different feature distributions of different modalities, the proposed model attempt to explore the interdependence between these modalities. As shown in Fig.~\ref{fig:zhangTT2}, there are two models trained by this team which one is trained using only IR as input and the other using both IR and Depth as inputs. Specially, a multi-stream architecture is designed with two sub-networks to perform multi-modal features fusion and the feature maps of two sub-networks is concatenated after a convolutional block. The final score is a weighted average of the results of two models. Their code is publicly available~\footnote{\url{https://github.com/ZhangTT-race/CVPR2020-MultiModal}}.

\noindent \textbf{Harvest.}~Different from other teams, they pay more attention to network structure, this team mainly explores data preprocessing and data augmentation to improve the generalization performance. Through experimental comparison, they found that IR modal data is more suitable for face anti-spoofing task. Therefore, in this multi-modal track, only the IR modal data participates in model training. Similar to the team Super, they first use the face detector to remove the background area outside the face. Concretely, they use a face detector to detect face ROI (Region of Interest) with RGB data, and then mapping theses ROIs to IR data to get the corresponding face position. Since only IR modal data is used, more sample augmentation strategies are used in network training to prevent overfitting. Such as the image is randomly divided into patches in online manner before sending to network. In addition, they tried some tricks including triplet loss with semi-hard negative mining, sample interpolation augmentation and label smoothing.

\begin{figure}
\begin{center}
\centering
\includegraphics[width=1.0\linewidth]{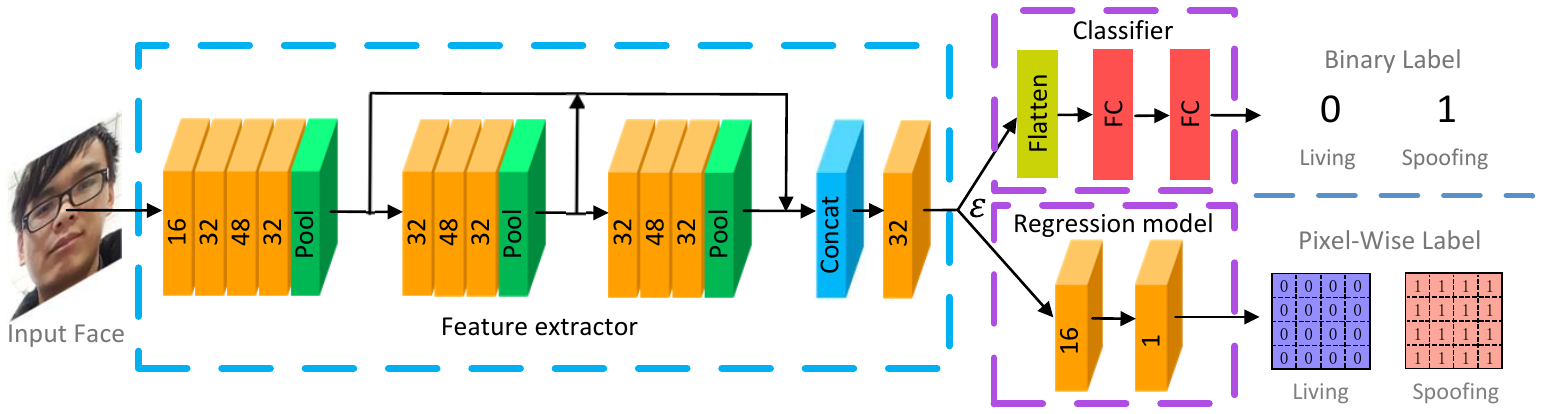}
\end{center}
\caption{The supervision and the network of Qyxqyx team. The orange cube is
convolution layer. The pixel-wise binary label in their experiment is resized into $32\times32$ resolution. The figure is provided by the Qyxqyx team and ranked $NO.7$ in multi-modal track.}
\label{fig:Qyxqyx}
\end{figure}

\noindent \textbf{Qyxqyx.}~Based on the work in~\cite{Liu2018Learning}, this team adds an additional binary classification supervision to promote the performance for multi-modal track. Specifically, the network structure is from~\cite{Liu2018Learning,qin2020learning} and the additional binary supervision is inspired by~\cite{DBLP:journals/corr/abs-1907-04047}. As shown in Fig.~\ref{fig:Qyxqyx}, taking the RGB modality as an example, the input samples are supervised by two loss functions which are a binary classification loss and a regression loss after passing through the feature network. Finally, the weighted sum of the binary output and the pixel-wise regression output as the final score. Their code is publicly available~\footnote{\url{https://github.com/qyxqyx/FAS_Chalearn_challenge}}.

\noindent \textbf{Skjack.}~The network structure is similar to team Super. They use ResNet-9~\cite{he2016deep} as backbone and fuse the RGB, Depth and IR features after the res-3 block, then a $1\times1$ convolution operation is used to compress the channel. Since there are no additional novel innovations, the team's algorithm did not perform well in this competition. Their code is publicly available~\footnote{\url{https://github.com/skJack/challange.git}}.

\section{Challenge Results}\label{sec4}
In this section, we first report the results of the participating teams from the perspective of both single-modal and multi-modal tracks, and then analyze the performances of the participants' methods. Finally, the shortcomings and limitations of these algorithms are pointed out.
\subsection{Challenge Results Report}
\subsubsection{Single-modal (RGB) Track}
Since the single-modal track only allows the use of RGB  data, the purpose is to evaluate the performance of the algorithms on a face anti-spoofing system with a VIS camera as the acquisition device. The final results of the $11$ participating teams are shown in Table~\ref{tab:single_results}, which includes the $3$ considered indicators (\eg, APCER, BPCER, ACER) on  three sub-protocols (\eg, $4\_1$, $4\_2$, $4\_3$). The final ranking is based on the average value of the ACER on three sub-protocols (smaller means better performance). At the same time, we report the thresholds for all algorithms to make decisions on real faces and attack samples. 
The thresholds of the top three teams are either very large (\ie, more than 0.9 for BOBO) or very small (\ie, 0.01 for Harvest), or have very different thresholds for different sub-protocols (\ie, 0.02 \vs 0.9 for VisionLabs). In addition, VisionLabs achieves the best results on APCER with the value of $2.72\%$, meaning that the algorithm can better classify attack samples correctly. Whilst, Wgqtmac's algorithm obtains the best results on the indicator of BPCER ($0.66\%$), indicating that it can better classify real face. Overall, the results of the first ten teams are better than the baseline method~\cite{li2020casiasurf} when ranking by ACER. The VisionLabs team achieved the first place with a clear advantage. 

\begin{table*}[]
\processtable{The results of Single-modal track. Avg$\pm$Std indicates the mean and variance operation and best results are shown in bold.\label{tab:single_results}}
{
\scalebox{0.82}{
\begin{tabular}{@{}clcccccccc@{}}
\toprule
Team Name & Method(keywords) & Prot. & Thre. & FP  & FN & APCER(\%) & BPCER(\%) & ACER(\%) & Rank \\ \midrule

\multirow{4}{*}{VisionLabs}
& \multirow{4}{*}{\begin{tabular}[c]{@{}l@{}}OpticalFlow,\\ RankPooling,\\ Data augment,\\ SimpleNet\end{tabular}}
& $4\_1$  & 0.02  & 4  & 21 & 0.22  & 5.25 & 2.74 & \multirow{4}{*}{1}\\ &
& $4\_2$  & 0.90  & 0  & 12 & 0.00  & 3.00 & 1.50   & \\  &
& $4\_3$  & 0.10 & 2 & 31 & 0.11 & 7.75  & 3.93  &  \\ &
& Avg$\pm$Std & 0.34$\pm$0.48 & 2$\pm$2      & 21$\pm$9  & \textbf{0.11$\pm$0.11} & 5.33$\pm$2.37   & \textbf{2.72$\pm$1.21} & \\ \midrule

\multirow{4}{*}{BOBO}
& \multirow{4}{*}{\begin{tabular}[c]{@{}l@{}}CDC, CDL, EDL,\\ Multi-level cell,\\ Attention moudle,\\ Depth supervision\end{tabular}}
& $4\_1$ & 0.95   & 201   & 10  & 11.17  & 2.5  & 6.83 & \multirow{4}{*}{2}  \\ &
& $4\_2$ & 0.99  & 120  & 8  & 6.67  & 2.0    & 4.33 &   \\ &
& $4\_3$  & 0.99   & 67  & 12  & 3.72   & 3.0  & 3.36 & \\ &
& Avg$\pm$Std & 0.97$\pm$0.02 & 129$\pm$67   & 10$\pm$2    & 7.18$\pm$3.74  & 2.50$\pm$0.50  & 4.84$\pm$1.79  &  \\ \midrule

\multirow{4}{*}{Harvest}
& \multirow{4}{*}{\begin{tabular}[c]{@{}l@{}}Motion cues,\\ relabelling live \\ sequence,\\ ResNet\end{tabular}}
& $4\_1$   & 0.01  & 31   & 48  & 1.72  & 12.0  & 6.86 & \multirow{4}{*}{3}  \\ &
& $4\_2$  & 0.01  & 116  & 51 & 6.44    & 12.75  & 9.6 &  \\ &
& $4\_3$  & 0.01  & 109  & 67 & 6.06    & 16.75  & 11.4  &  \\ &
& Avg$\pm$Std & 0.01$\pm$0.00 & 85$\pm$47    & 55$\pm$10  & 4.74$\pm$2.62 & 13.83$\pm$2.55  & 9.28$\pm$2.28  &  \\ \midrule

\multirow{4}{*}{ZhangTT}
& \multirow{4}{*}{\begin{tabular}[c]{@{}l@{}}Quality tensor,\\ Time tensor,\\ Data \\ preprocessing\end{tabular}}
& $4\_1$  & 0.9  & 103   & 74   & 5.72 & 18.5  & 12.11  & \multirow{4}{*}{4}  \\ &
& $4\_2$  & 0.9   & 132  & 45  & 7.33 & 11.25 & 9.29  &  \\ &
& $4\_3$  & 0.9   & 57   & 108 & 3.17 & 27.0  & 15.08 &  \\ &
& Avg$\pm$Std & 0.9  & 97$\pm$37  & 75$\pm$31 & 5.40$\pm$2.10 & 18.91$\pm$7.88  & 12.16$\pm$2.89  &   \\ \midrule

\multirow{4}{*}{\begin{tabular}[c]{@{}c@{}}Newland-\\ tianyan\end{tabular}}
& \multirow{4}{*}{\begin{tabular}[c]{@{}l@{}}Data augment,\\ Temporal feature,\\ Neighborhood \\ mean\end{tabular}}
& $4\_1$  & 0.77  & 34   & 117   & 1.89   & 29.25   & 15.57  & \multirow{4}{*}{5}  \\ &
& $4\_2$   & 0.7 & 513 & 11 & 28.5 & 2.75  & 15.62 & \\ &
& $4\_3$   & 0.55  & 299  & 6   & 16.61  & 1.5  & 9.06  &  \\ &
& Avg$\pm$Std & 0.67$\pm$0.11 & 282$\pm$239  & 44$\pm$62   & 15.66$\pm$13.33& 11.16$\pm$15.67   & 13.41$\pm$3.77  &  \\ \midrule

\multirow{4}{*}{Dopamine}
& \multirow{4}{*}{\begin{tabular}[c]{@{}l@{}}ID information,\\ Multi-task,\\ Score fusion,\\ Resnet100\end{tabular}}
& $4\_1$ & 0.02  & 325  & 6  & 18.06  & 1.5  & 9.78  & \multirow{4}{*}{6}  \\ &
& $4\_2$  & 0.22  & 367  & 24  & 20.39  & 6.0 & 13.19 &  \\ &
& $4\_3$  & 0.01  & 636  & 0 & 35.33  & 0.0  & 17.67  & \\ &
& Avg$\pm$Std & 0.07$\pm$0.11 & 442$\pm$168  & 10$\pm$12   & 24.59$\pm$9.37 & 2.50$\pm$3.12 & 13.54$\pm$3.95 & \\ \midrule

\multirow{4}{*}{IecLab}
& \multirow{4}{*}{\begin{tabular}[c]{@{}l@{}}3D ResNet,\\ Fueature fusion,\\ Softmax\end{tabular}}
& $4\_1$  & 0.33  & 696  & 21  & 38.67  & 5.25 & 21.96    & \multirow{4}{*}{7}  \\ &
& $4\_2$  & 0.45  & 606 & 26 & 33.67 & 6.5 & 20.08 &   \\ &
& $4\_3$  & 0.45  & 489 & 26 & 27.17 & 6.5 & 16.83 &   \\ &
& Avg$\pm$Std & 0.40$\pm$0.07 & 597$\pm$103  & 24$\pm$2  & 33.16$\pm$5.76   & 6.08$\pm$0.72  & 19.62$\pm$2.59  &  \\ \midrule

\multirow{4}{*}{\begin{tabular}[c]{@{}c@{}}Chunghwa-\\ Telecom\end{tabular}}
& \multirow{4}{*}{\begin{tabular}[c]{@{}l@{}}Subsequence \\ feature,\\ Local feature,\\ MIMAMO-Net\end{tabular}}
& $4\_1$  & 0.87  & 538  & 44  & 29.89    & 11.0  & 20.44  & \multirow{4}{*}{8}  \\ &
& $4\_2$ & 0.93  & 352  & 113  & 19.56 & 28.25  & 23.9  & \\ &
& $4\_3$ & 0.79  & 442  & 71   & 24.56 & 17.75  & 21.15 & \\ &
& Avg$\pm$Std & 0.86$\pm$0.06 & 444$\pm$93   & 76$\pm$34  & 24.66$\pm$5.16                          & 19.00$\pm$8.69  & 21.83$\pm$1.82 & \\ \midrule

\multirow{4}{*}{Wgqtmac}
& \multirow{4}{*}{\begin{tabular}[c]{@{}l@{}}ResNet18,\\ Warmup strategy,\\ Softmax\end{tabular}}
& $4\_1$  & 0.85 & 1098  & 1  & 61.0  & 0.25                                    & 30.62  & \multirow{4}{*}{9}  \\ &
& $4\_2$ & 1.0 & 570  & 7   & 31.67 & 1.75 & 16.71 &  \\ &
& $4\_3$ & 0.56  & 1117 & 0 & 62.06 & 0.0    & 31.03 &  \\ &
& Avg$\pm$Std & 0.80$\pm$0.22 & 928$\pm$310  & 2$\pm$3  & 51.57$\pm$17.24   & \textbf{0.66$\pm$0.94} & 26.12$\pm$8.15 &  \\ \midrule

\multirow{4}{*}{Hulking}
& \multirow{4}{*}{\begin{tabular}[c]{@{}l@{}}PipeNet,\\ Softamx\end{tabular}}
& $4\_1$  & 0.81  & 635  & 138  & 35.28  & 34.5  & 34.89  & \multirow{4}{*}{10} \\ &
& $4\_2$  & 0.82  & 1027 & 37   & 57.06  & 9.25  & 33.15  &  \\ &
& $4\_3$  & 0.67  & 768  & 59   & 42.67  & 14.75 & 28.71  &  \\ &
& Avg$\pm$Std & 0.76$\pm$0.08 & 810$\pm$199  & 78$\pm$53  & 45.00$\pm$11.07 & 19.50$\pm$13.27  & 32.25$\pm$3.18  & \\ \midrule

\multirow{4}{*}{Dqiu}
& \multirow{4}{*}{\begin{tabular}[c]{@{}l@{}}ResNet50,\\ Softmax\end{tabular}}
& $4\_1$  & 1.0  & 1316  & 142 & 73.11 & 35.5 & 54.31 & \multirow{4}{*}{11} \\ &
& $4\_2$  & 1.0  & 567   & 60  & 31.5  & 15.0 & 23.25 & \\ &
& $4\_3$  & 1.0  & 664   & 146 & 36.89 & 36.5 & 36.69 & \\ &
& Avg$\pm$Std & 1.00$\pm$0.00 & 849$\pm$407  & 116$\pm$48 & 47.16$\pm$22.62
& 29.00$\pm$12.13 & 38.08$\pm$15.57 & \\ \midrule

\multirow{4}{*}{Baseline}
& \multirow{4}{*}{\begin{tabular}[c]{@{}l@{}}Static and \\ Dynamic\\ features,\\ RankPooling\end{tabular}}
& $4\_1$  & 1.0  & 1331  & 7  & 73.94  & 1.75 & 37.85 & \multirow{4}{*}{*}  \\ &
& $4\_2$  & 1.0  & 1379  & 27 & 76.61  & 6.75 & 41.68 & \\ &
& $4\_3$  & 1.0  & 836   & 57 & 46.44  & 14.25 & 30.35 &  \\ &
& Avg$\pm$Std & 1.00$\pm$0.00 & 1182$\pm$300 & 30$\pm$25  & 65.66$\pm$16.70     & 7.58$\pm$6.29 & 36.62$\pm$5.76 &  \\ \bottomrule
\end{tabular}
}
}{}
\end{table*}

\subsubsection{Multi-modal}
The Multi-modal Track allows the participating teams to use all the modal data. The purpose is to evaluate the performance of the algorithms on anti-spoofing systems equipped with multi-optic cameras, such as the Intel RealSense or Microsoft Kinect sensor. The results of the eight participating teams in the final stage are shown in Table~\ref{tab:multi_results}. BOBO team's algorithm gets first place performance, such as $APCER=1.05\%$, $BPCER=1.00\%$, and $ACER=1.02\%$. While the team of Super ranks second with a slight disadvantage, such as $ACER=1.68\%$. It is worth noting that Newland-tianyan's algorithm achieves the best results on the APCER indicator with the value of $0.24\%$. Similar to the conclusion of the single-modal track, most of the participating teams have relatively large thresholds which are calculated on the validation set, especially the Super and Newland-tianyan teams with the value of $1.0$ on three sub-protocols, indicating that these algorithms treat the face anti-spoofing task as an anomaly detection. In addition, we can find that the ACER values of the top four teams are $1.02\%$, $1.68\%$, $2.21\%$, and $2.28\%$, which are better than the ACER of the first place of the single-modal track, such as $2.72\%$ for the team of VisionLabs. It shows the necessity of our multi-modal track in improving accuracy in face anti-spoofing task.
\begin{table*}[]
\processtable{The results of Multi-modal track. Avg$\pm$Std indicates the mean and variance operation and best results are shown in bold.\label{tab:multi_results}}
{
\scalebox{0.82}{
\begin{tabular}{clcccccccc}
\toprule
Team Name   & Method(keywords)  & Prot.       & Thre.         & FP             & FN    & APCER(\%)  & BPCER(\%)     & ACER(\%)      & Rank      \\ \midrule
\multirow{4}{*}{BOBO}
& \multirow{4}{*}{\begin{tabular}[c]{@{}l@{}}CDC, CDL, EDL,\\ Feature fusion,\\ Score fusion,\\ Depth supervision\end{tabular}}
& $4\_1$      & 0.98          & 6              & 2         & 0.33                   & 0.5                    & 0.42                   & \multirow{4}{*}{1} \\ &
& $4\_2$      & 0.95          & 25             & 3         & 1.39                   & 0.75                   & 1.07                   &                    \\ &
& $4\_3$      & 0.94          & 26             & 7         & 1.44                   & 1.75                   & 1.6                    &                    \\ &
& Avg$\pm$Std & 0.95$\pm$0.02 & 19$\pm$11      & 4$\pm$2   & 1.05$\pm$0.62          & \textbf{1.00$\pm$0.66} & \textbf{1.02$\pm$0.59} &                    \\ \midrule

\multirow{4}{*}{Super}
& \multirow{4}{*}{\begin{tabular}[c]{@{}l@{}}Data preprocessing,\\ Dimension reduction,\\ SE fusion,\\ Score fusion\end{tabular}}
& $4\_1$      & 1.0           & 9              & 11        & 0.5                    & 2.75                   & 1.62                   & \multirow{4}{*}{2} \\  &                                         & $4\_2$      & 1.0           & 5              & 17        & 0.28                   & 4.25                   & 2.26                   &                    \\  &
& $4\_3$      & 1.0           & 20             & 5         & 1.11                   & 1.25                   & 1.18                   &                    \\  &                                         & Avg$\pm$Std & 1.0$\pm$0.00  & 11.33$\pm$7.76 & 11$\pm$6  & 0.62$\pm$0.43          & 2.75$\pm$1.50          & 1.68$\pm$0.54          &                    \\ \midrule

\multirow{4}{*}{Hulking}
& \multirow{4}{*}{\begin{tabular}[c]{@{}l@{}}PipeNet,\\ SENet-154,\\ Selective Modal Pipeline,\\ Limited Frame Vote\end{tabular}}
& $4\_1$      & 0.96          & 31             & 0         & 1.72                   & 0.0                    & 0.86                   & \multirow{4}{*}{3} \\  &
& $4\_2$      & 1.0           & 99             & 5         & 5.5                    & 1.25                   & 3.37                   &                    \\  &                                         & $4\_3$      & 1.0           & 46             & 9         & 2.56                   & 2.25                   & 2.4                    &                    \\  &
& Avg$\pm$Std & 0.98$\pm$0.02 & 58$\pm$35      & 4$\pm$4   & 3.25$\pm$1.98          & 1.16$\pm$1.12          & 2.21$\pm$1.26          &                    \\ \midrule

\multirow{4}{*}{\begin{tabular}[c]{@{}c@{}}Newland-\\ tianyan\end{tabular}}
& \multirow{4}{*}{\begin{tabular}[c]{@{}l@{}}Resnet9,\\ Data preprocessing,\\ Neighborhood mean,\\ Data augment\end{tabular}}
& $4\_1$      & 1.0           & 0              & 3         & 0.0                    & 0.75                   & 0.37                   & \multirow{4}{*}{4} \\  &
& $4\_2$      & 1.0           & 4              & 26        & 0.22                   & 6.5                    & 3.36                   &                    \\  &
& $4\_3$      & 1.0           & 9              & 23        & 0.5                    & 5.75                   & 3.12                   &                    \\  &
& Avg$\pm$Std & 1.00$\pm$0.00 & 4$\pm$4        & 17$\pm$12 & \textbf{0.24$\pm$0.25} & 4.33$\pm$3.12          & 2.28$\pm$1.66          &                    \\ \midrule

\multirow{4}{*}{ZhangTT}
& \multirow{4}{*}{\begin{tabular}[c]{@{}l@{}}ID Net,\\ Feature fusion,\\ Score fusion\end{tabular}}
& $4\_1$      & 0.94          & 0              & 19        & 0.0                    & 4.75                   & 2.37                   & \multirow{4}{*}{5} \\  &
& $4\_2$      & 0.9           & 66             & 34        & 3.67                   & 8.5                    & 6.08                   &                    \\  &                                         & $4\_3$      & 0.79          & 102            & 0         & 5.67                   & 0.0                    & 2.83                   &                    \\  &                                         & Avg$\pm$Std & 0.87$\pm$0.07 & 56$\pm$51      & 17$\pm$17 & 3.11$\pm$2.87          & 4.41$\pm$4.25          & 3.76$\pm$2.02          &                    \\ \midrule

\multirow{4}{*}{Harvest}
& \multirow{4}{*}{\begin{tabular}[c]{@{}l@{}}Data preprocessing,\\ Data augment,\\ Only IR,\\ Semi-hard negative mining\end{tabular}}
& $4\_1$      & 0.87          & 13             & 4         & 0.72                   & 1.0                    & 0.86                   & \multirow{4}{*}{6} \\  &
& $4\_2$      & 0.93          & 180            & 28        & 10.0                   & 7.0                    & 8.5                    &                    \\  &                                         & $4\_3$      & 0.96          & 119            & 8         & 6.61                   & 2.0                    & 4.31                   &                    \\  &                                         & Avg$\pm$Std & 0.92$\pm$0.04 & 104$\pm$84     & 13$\pm$12 & 5.77$\pm$4.69          & 3.33$\pm$3.21          & 4.55$\pm$3.82          &                    \\ \midrule

\multirow{4}{*}{Qyxqyx}
& \multirow{4}{*}{\begin{tabular}[c]{@{}l@{}}Binary supervision,\\ Pixel-wise regression,\\ Score fusion\end{tabular}}
& $4\_1$      & 0.98          & 1              & 53        & 0.06                   & 13.25                  & 6.65                   & \multirow{4}{*}{7} \\  &
& $4\_2$      & 0.98          & 19             & 8         & 1.06                   & 2.0                    & 1.53                   &                    \\  &
& $4\_3$      & 0.89          & 257            & 19        & 14.28                  & 4.75                   & 9.51                   &                    \\  &
& Avg$\pm$Std & 0.95$\pm$0.05 & 92$\pm$142     & 26$\pm$23 & 5.12$\pm$7.93          & 6.66$\pm$5.86          & 5.89$\pm$4.04          &                    \\ \midrule

\multirow{4}{*}{Skjack}
& \multirow{4}{*}{\begin{tabular}[c]{@{}l@{}}Resnet9,\\ Softmax\end{tabular}}
& $4\_1$      & 0.0           & 1371           & 2         & 76.17                  & 0.5                    & 38.33                  & \multirow{4}{*}{8} \\  &
& $4\_2$      & 0.01          & 1155           & 46        & 64.17                  & 11.5                   & 37.83                  &                    \\  &                                         & $4\_3$      & 0.0           & 511            & 93        & 28.39                  & 23.25                  & 25.82                  &                    \\  &
& Avg$\pm$Std & 0.00$\pm$0.00 & 1012$\pm$447   & 47$\pm$45 & 56.24$\pm$24.85        & 11.75$\pm$11.37        & 33.99$\pm$7.08         &                    \\ \midrule

\multirow{4}{*}{Baseline}
& \multirow{4}{*}{\begin{tabular}[c]{@{}l@{}}SD-Net,\\ A shared branch,\\ PSMM-Net,\\ Fusion fusion,\end{tabular}}
& $4\_1$      & 1.0           & 413            & 109       & 22.94                  & 27.25                  & 25.1                   & \multirow{4}{*}{*} \\  &                                         & $4\_2$      & 0.17          & 1340           & 23        & 74.44                  & 5.75                   & 40.1                   &                    \\  &
& $4\_3$      & 0.02          & 864            & 55        & 48.0                   & 13.75                  & 30.87                  &                    \\  &                                         & Avg$\pm$Std & 0.39$\pm$0.52 & 872$\pm$463    & 62$\pm$43 & 48.46$\pm$25.75        & 15.58$\pm$10.86        & 32.02$\pm$7.56         &                    \\ \bottomrule
\end{tabular}
}
}{}
\end{table*}

\subsection{Challenge Results Analysis}
In this section we  analyze the advantages and disadvantages of the algorithm performance of each participating team in detail according to different tracks.
\subsubsection{Single-modal}
As shown in Table~\ref{tab:PandS}, the testing subset introduces two unknown target variations simultaneously, such as the different ethnicities and attack types in training and testing subsets, which pose the huge challenge for participating teams. However, most teams achieved relatively good results in the final stage compared to baseline, especially the top three teams get ACER values below $10\%$. It is worth mentioning that different algorithms have their own unique advantages, even if the final ranking is relatively backward. Such as the value of BPCER of Wgqtmac'team is $0.66\%$, meaning about 1 real sample from 100 real face will be treated as fake ones. While, APCER=$0.11\%$ for the team of VisionLabs indicates about 1 fake samples from 1000 attackers will be treated as real ones.

\begin{figure*}[t]
	\begin{center}
	\includegraphics[width=1.0\linewidth]{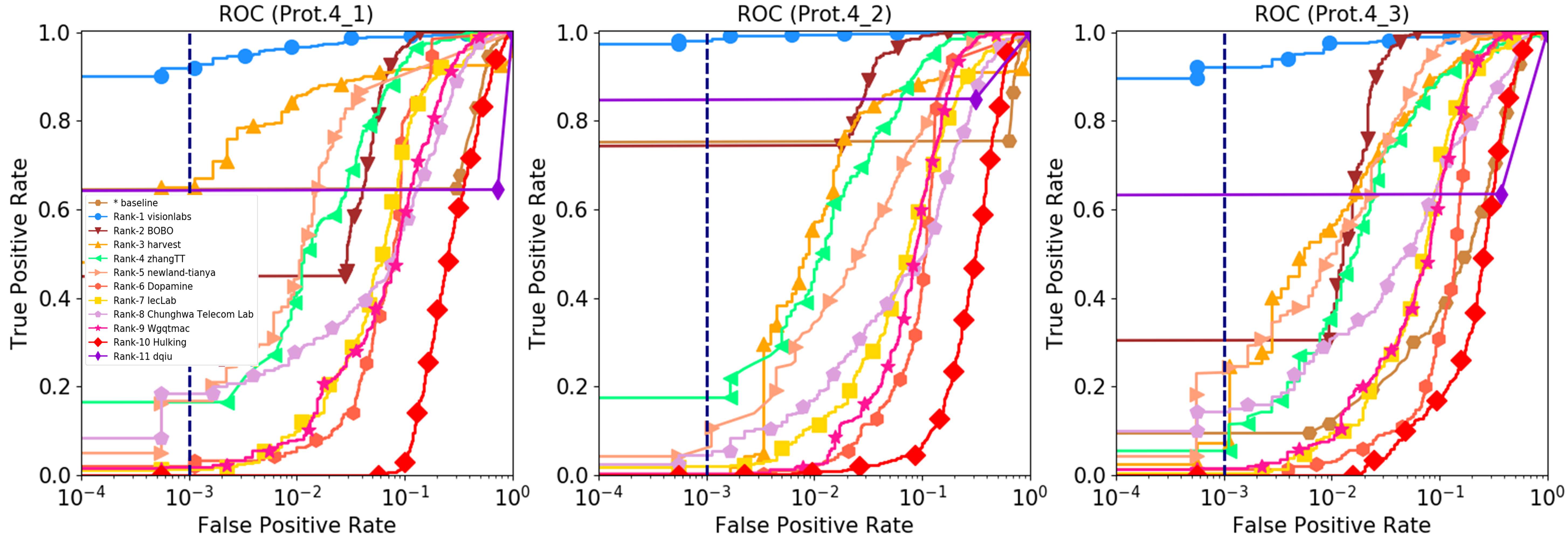}
	\end{center}
	\caption{The ROC of $12$ teams in single-modal track. From left to right are the ROCs on protocol 4\_1, 4\_2 and 4\_3.}
	\label{fig:sroc}
\end{figure*}

To fully compare the stability of the participating team's algorithms, similar to~\cite{DBLP:conf/cvpr/abs-1812-00408}, we introduce the receiver operating characteristic (ROC) curve in this challenge which can be used to select a suitable trade off threshold between false positive rate (FPR) and true positive rate (TPR) according to the requirements of a given real application. As shown in Fig.~\ref{fig:sroc}, the results of the top one team (VisionLabs) on both three sub-protocols are clearly superior to other teams, revealing that using optical flow method to convert RGB modal data to other sample spaces can effectively improve the generalization performance of the algorithm to deal with different unknown factors. However, the TPR value of the remaining teams decreased rapidly as the FPR reduced (\textit{e.g.}, TPR@FPR=$10^{-3}$ values of these teams are almost zero). 
In addition, we can find that although the performance of ACER for Harvest team is worse than that of the BOBO team, the performance of the TPR@FPR=$10^{-3}$ is significantly better than the BOBO team. It is mainly because the false positive (FP) and false negative (FN) samples of the Harvest team are relatively close (see from Table~\ref{tab:single_results}).

\begin{figure}[t]
\begin{center}
\includegraphics[width=1.0\linewidth]{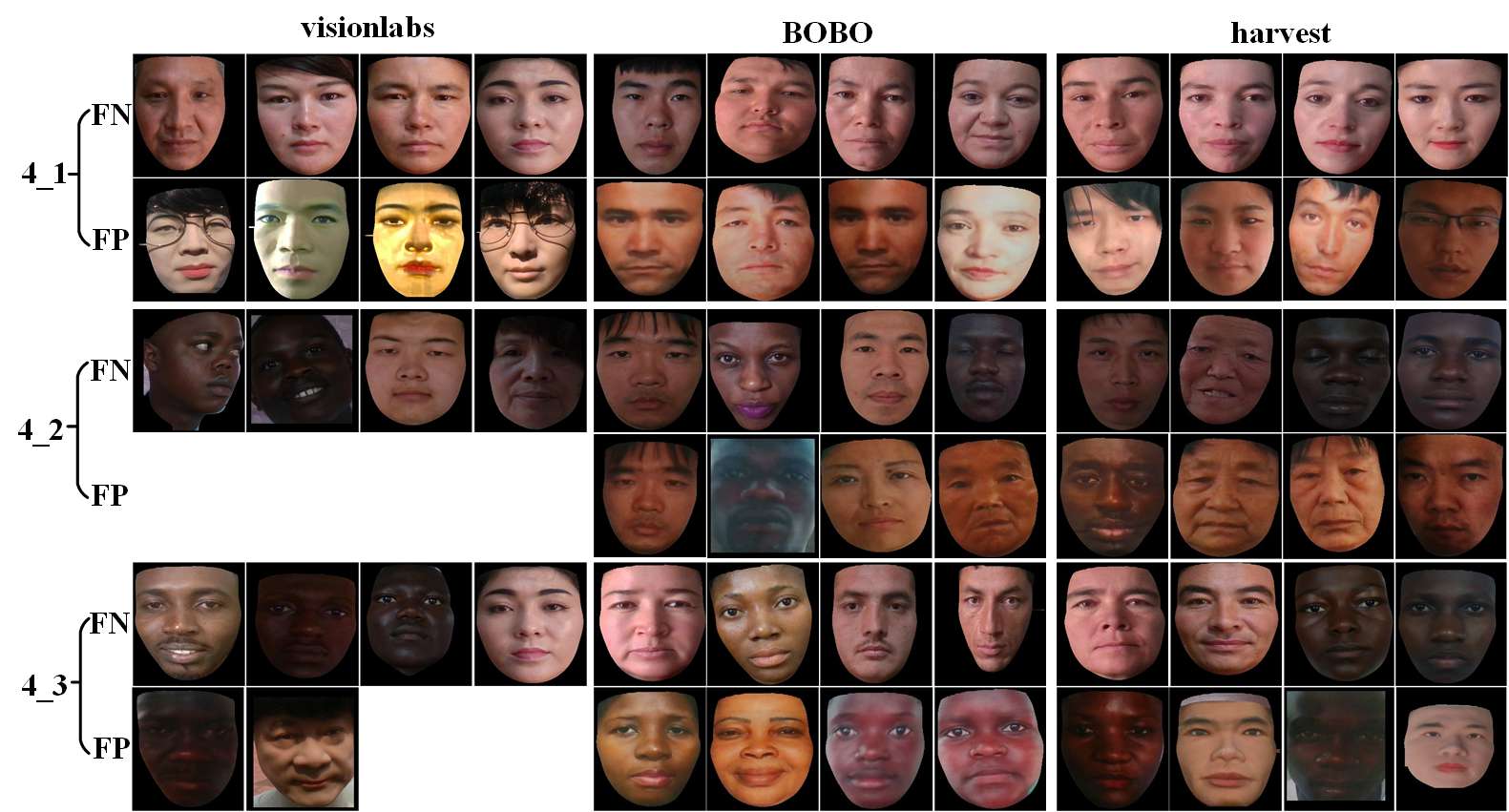}
\end{center}
\caption{The mismatched samples of top three teams in single-modal track. FN and FP indicate false negative and false positive respectively.}
\label{fig:sfnfp}
\end{figure}

Finally, for the top three teams, we randomly selected some mismatched samples as shown in Fig.~\ref{fig:sfnfp}. We can see that most of the FN samples of the VisionLabs team are real faces with large motion amplitude, while the most of FP samples are 3D print attacks, indicating that the team's algorithm has correctly classified almost all 2D attack samples. In addition, due to the challenging nature of our competition dataset, such as it is difficult to distinguish the real face from attack samples without the label, the BOBO team and the Harvest team did not make correct decisions on some difficult samples.

\subsubsection{Multi-modal}
\begin{figure*}[t]
	\begin{center}
	\includegraphics[width=1.0\linewidth]{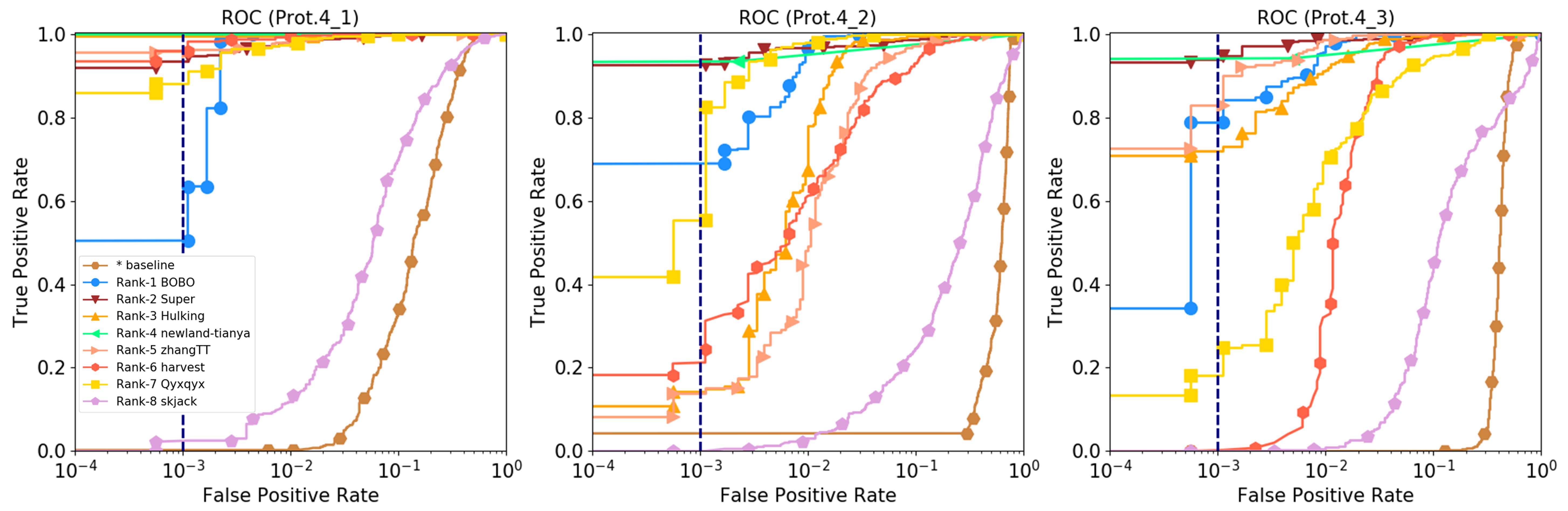}
	\end{center}
	\caption{The ROC of $9$ teams in multi-modal track. From left to right are the ROCs on protocol 4\_1, 4\_2 and 4\_3.}
	\label{fig:mroc}
\end{figure*}

\begin{figure}[t]
	\begin{center}
	\includegraphics[width=1.0\linewidth]{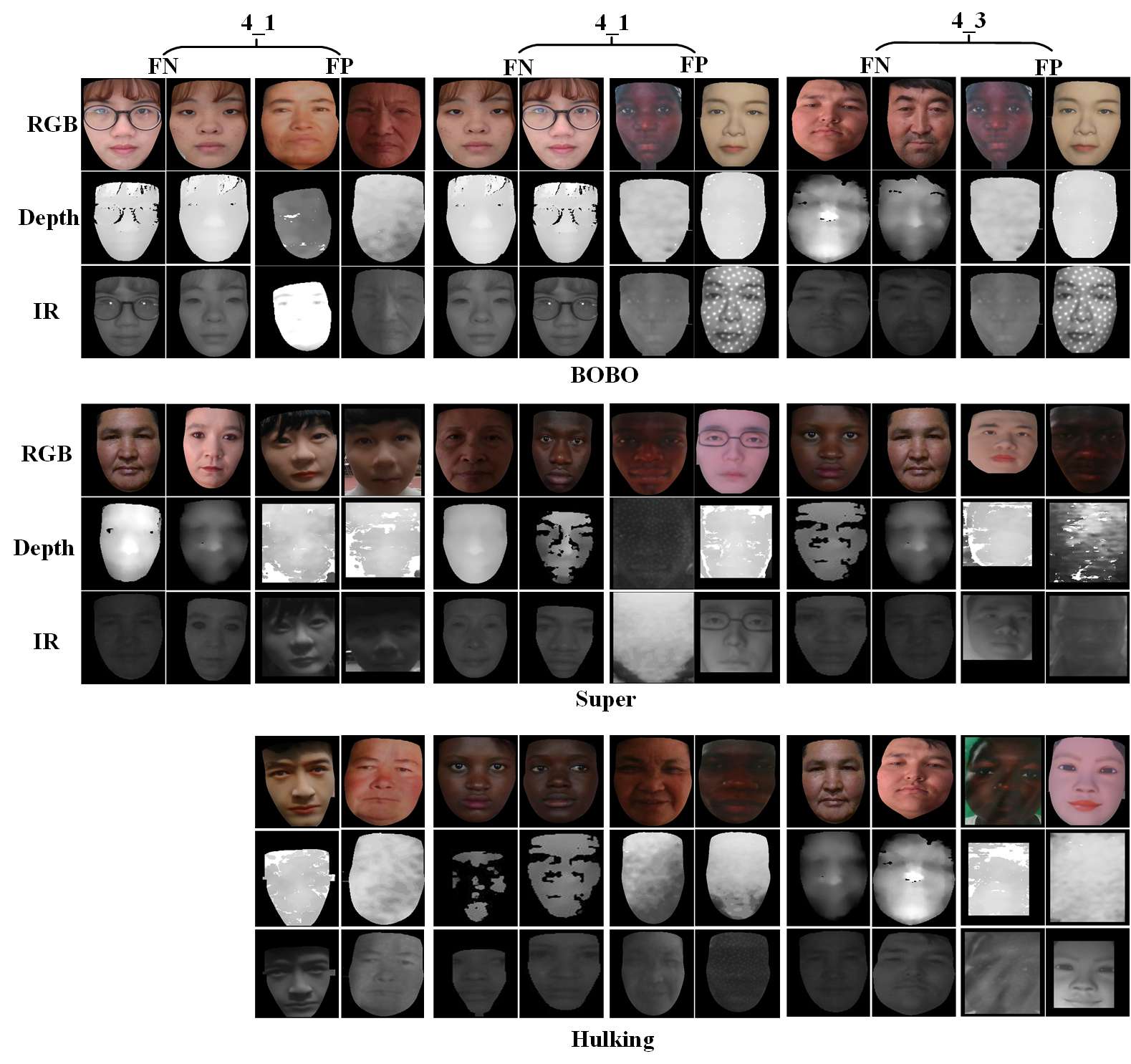}
	\end{center}
	\caption{The mismatched samples of top three teams in multi-modal track. FN and FP indicate false negative and false positive respectively.}
	\label{fig:mfnfp}
\end{figure}

From the Table~\ref{tab:multi_results}, we can find that the ACER values of the top 7 teams are relatively close, and the top 4 teams are better than VisionLabs (ACER=$2.72\%$) in single-modal track. It indicates that the complementary information between multi-modal datasets can improve the accuracy of the face anti-spoofing algorithm. Although newland-tianyan ranked fourth in ACER, they achieved the best results on the APCER indicator (\eg, APCER=$0.24\%$). It means the smallest number of FP samples among all teams. In addition, from the Table~\ref{tab:multi_results} and Fig.~\ref{fig:mroc}, we can find that although the ACER values of the top two algorithms are relatively close, the stability of the Super team is better than the BOBO, such as the values of TPR@FPR=$10^{-3}$ for Super and newland-tianyan are better than BOBO on both three sub-protocols. Finally, we can find from the Fig.~\ref{fig:mfnfp} that the FP samples of the top three teams contain many 3D print attacks, indicating that their algorithms are vulnerable to 3D face attacks.

\section{Open Issues and Opportunities}\label{sec5}
In this section, we will first summarize some common issues that appear in this challenge, then analyze some of the causes that result to the problems, and put forward some feasible solutions to alleviate these problems in combination with practical applications. Finally we formulate the future work based on the CASIA-SURF CeFA dataset.

\subsection{Critical Issues and Breakthrough Point}
From Tables~\ref{tab:single_results} and~\ref{tab:multi_results} of the competition results, we can find that the threshold for both single-modal and multi-modal track is generally high. This is while the meaning of the threshold in our challenge is the minimum probability that a sample will be classified as a real face. For instance, the thresholds on three sub-protocols reach to 1 for the team of dqiu in single-modal track and the top-ranked teams (Super, Hulking and newland-tianyan) in multi-modal track. These over-confidence problems mean that some attack samples will be judged as real faces with high probability, which is unreasonable in practical applications. We analyze the following three reasons responsible for this problem: (1) Caused by the task itself. The nature of the face anti-spoofing task is a binary classification task. If the sample scale is small and lacks diversity, it can easily lead to extreme thresholds. This phenomenon is also found in other binary classification tasks, such as face detection. (2) Caused by different collection environments for positive (real face) and negative samples (spoof). For example, the attack samples of the same subject are collected under multiple lighting conditions, while real face is collected only in indoor environments. (3) Caused by the lack of generalization performance when the algorithm faces unknown attack types and ethnicities. According to the characteristics of the testing protocol that contains 2 unknown variables (\ie, cross-PAIs and cross-ethnicity) in training and testing phases, some teams design networks and loss functions pay more attention to the motion information of real face and replay attack in the training phase, and treat any unseen static-samples (including spoofs and real faces) in the testing phase as abnormal information (spoofs), resulting in poor generalization ability in cross-PAIs. Other teams have subtracted different neighborhood mean values according to different ethnicities to alleviate the interference caused by skin color differences. However, in the face of unknown ethnic samples, the inability to subtract the appropriate neighborhood mean causes classification errors. In summary, poor generalization performance (\ie, unable to correctly classify unknown real samples and attack types) causes the classification threshold to be too large or too small. To alleviate this problem, we propose feasible solutions from the three aspects of data collection, training strategy and algorithm design. CASIA-SURF CeFA is the largest up to date face anti-spoofing dataset and contains various attack types and attack environments, such as the attack types include print attacks and replay attacks under multiple lighting conditions. 
However, the diversity of the device and environment for collecting real face samples is limited.
It inevitably brings the problem of sample imbalance. Therefore, the CASIA-SURF CeFA dataset should consider supplementing some real samples including acquisition equipment and shooting environment. Whilst, an effective training strategy is to balance the positive and negative proportions of samples in each batch during the training process. Finally, a binary cross-entropy loss might discover arbitrary cues, such as spot or screen bezel of the spoof medium, that are not the faithful spoof patterns. Therefore, the supervision should be designed from the essential differences between live and spoof faces, such as the rPPG signals (\textit{i.e.}, heart pulse signal) which can reflect human physiological signs.

\subsection{Future Work and Opportunities}
Face anti-spoofing based on multi-modal datasets attract increasing research interests. However, the gap exploration between sensing patterns of different face modalities remains an open research problem in face anti-spoofing. Some previous works~\cite{li2020casiasurf,zhang2019casiasurf} have been verified the existence of performance deviations of the SOTA algorithms in different face modalities. At the same time, they designed a testing protocol to measure the degree of modal bias, such as the Protocol 3 in CASIA-SURF CeFA~\cite{li2020casiasurf}. Similar to heterogeneous face recognition (\eg, NIR-VIS~\cite{Yi2007Face,Lei2009Coupled,lezama2016afraid}), which refers to matching faces across different modalities (or sensing patterns), we cast the face anti-spoofing task as a heterogeneous face matching problem. In this way, the discrimination information of other modal samples can be used to assist the learning of RGB modal data. And after the model is trained, there is no need to load other modal samples during the testing phase.

Since the existing datasets for training and verification are collected in VIS spectrum, the use of samples of additional modalities (\eg, Depth or IR) to assist the learning of RGB modal data while without extra modalities in testing phase is interesting in the  practical applications. 
On the other hand, CASIA-SURF~\cite{DBLP:conf/cvpr/abs-1812-00408} and CASIA-SURF CeFA~\cite{li2020casiasurf} are multi-modal face anti-spoofing datasets and each sample contains 3 paired modalities, which may provide us with the possibility to study heterogeneous face anti-spoofing.

\section{Conclusion}\label{sec6}
We organized the \emph{Chalearn Face Anti-spoofing Attack Detection Challenge at CVPR2020} based on the CASIA-SURF CeFA dataset with two tracks and running on the CodaLab platform. Both tracks attracted $340$ teams in the development stage, and finally $11$ and $8$ teams have submitted their codes in the single-modal and multi-modal face anti-spoofing recognition challenges, respectively.  
We described the associated dataset, and the challenge protocol including evaluation metrics. We reviewed in detail the proposed solutions and reported the challenge results. Compared with the baseline method, the best performances from participants under the ACER value are from 36.62 to 2.72, and 32.02 to 1.02 for the single-modal and multi-modal challenges, respectively. We analyzed the results of the challenge, pointing out the critical issues in FAD task and presenting the shortcomings of the existing algorithms. Future lines of research in the field have been also discussed.

\section{Acknowledgments}\label{sec7}
This work has been partially supported by Science and Technology Development Fund of Macau (Grant No. 0025/2018/A1), by the Chinese National Natural Science Foundation Projects $\#$61876179, $\#$61872367, the Spanish project TIN2016-74946-P (MINECO/FEDER, UE) and CERCA Programme / Generalitat de Catalunya, and by ICREA under the ICREA Academia programme. We acknowledge Surfing Technology Beijing co., Ltd (www.surfing.ai) to provide us this high quality dataset. We also acknowledge the support of NVIDIA Corporation with the donation of the GPU used for this research. Finally, we thank all participating teams for their participation and contributions, and special thanks to VisionLabs, BOBO, Harvset, ZhangTT, Newland-tianyan, Dopamine, Hulking, Super, Qyxqyx for their guidance in drawing figure.

\bibliographystyle{IEEEtran}
\bibliography{IEEEabrv,sample}

\begin{thebibliography}{10}
\providecommand{\url}[1]{#1}
\csname url@samestyle\endcsname
\providecommand{\newblock}{\relax}
\providecommand{\bibinfo}[2]{#2}
\providecommand{\BIBentrySTDinterwordspacing}{\spaceskip=0pt\relax}
\providecommand{\BIBentryALTinterwordstretchfactor}{4}
\providecommand{\BIBentryALTinterwordspacing}{\spaceskip=\fontdimen2\font plus
\BIBentryALTinterwordstretchfactor\fontdimen3\font minus
  \fontdimen4\font\relax}
\providecommand{\BIBforeignlanguage}[2]{{%
\expandafter\ifx\csname l@#1\endcsname\relax
\typeout{** WARNING: IEEEtran.bst: No hyphenation pattern has been}%
\typeout{** loaded for the language `#1'. Using the pattern for}%
\typeout{** the default language instead.}%
\else
\language=\csname l@#1\endcsname
\fi
#2}}
\providecommand{\BIBdecl}{\relax}
\BIBdecl

\bibitem{Pan2007Eyeblink}
G.~Pan, L.~Sun, Z.~Wu, and S.~Lao, ``Eyeblink-based anti-spoofing in face
  recognition from a generic webcamera,'' in \emph{ICCV}, 2007.

\bibitem{wang2009face}
L.~Wang, X.~Ding, and C.~Fang, ``Face live detection method based on
  physiological motion analysis,'' \emph{TST}, 2009.

\bibitem{kollreider2008verifying}
K.~Kollreider, H.~Fronthaler, and J.~Bigun, ``Verifying liveness by multiple
  experts in face biometrics,'' in \emph{CVPR Workshops}, 2008.

\bibitem{Bharadwaj2013Computationally}
S.~Bharadwaj, T.~I. Dhamecha, M.~Vatsa, and R.~Singh, ``Computationally
  efficient face spoofing detection with motion magnification,'' in
  \emph{CVPR}, 2013.

\bibitem{Pan2011Monocular}
G.~Pan, L.~Sun, Z.~Wu, and Y.~Wang, ``Monocular camera-based face liveness
  detection by combining eyeblink and scene context,'' \emph{TCS}, 2011.

\bibitem{Komulainen2014Context}
J.~Komulainen, A.~Hadid, and M.~Pietikainen, ``Context based face
  anti-spoofing,'' in \emph{BTAS}, 2013.

\bibitem{chingovska2012effectiveness}
I.~Chingovska, A.~Anjos, and S.~Marcel, ``On the effectiveness of local binary
  patterns in face anti-spoofing,'' in \emph{BIOSIG}, 2012.

\bibitem{Yang2013Face}
J.~Yang, Z.~Lei, S.~Liao, and S.~Z. Li, ``Face liveness detection with
  component dependent descriptor.'' in \emph{ICB}, 2013.

\bibitem{Maatta2012Face}
J.~Maatta, A.~Hadid, and M.~Pietikainen, ``Face spoofing detection from single
  images using texture and local shape analysis,'' \emph{IET biometrics}, 2012.

\bibitem{schwartz2011face}
W.~R. Schwartz, A.~Rocha, and H.~Pedrini, ``Face spoofing detection through
  partial least squares and low-level descriptors,'' in \emph{IJCB}, 2011.

\bibitem{feng2016integration}
L.~Feng, L.-M. Po, Y.~Li, X.~Xu, F.~Yuan, T.~C.-H. Cheung, and K.-W. Cheung,
  ``Integration of image quality and motion cues for face anti-spoofing: A
  neural network approach,'' \emph{JVCIR}, 2016.

\bibitem{li2016original}
L.~Li, X.~Feng, Z.~Boulkenafet, Z.~Xia, M.~Li, and A.~Hadid, ``An original face
  anti-spoofing approach using partial convolutional neural network,'' in
  \emph{IPTA}, 2016.

\bibitem{Patel2016Secure}
K.~Patel, H.~Han, and A.~K. Jain, ``Secure face unlock: Spoof detection on
  smartphones,'' \emph{TIFS}, 2016.

\bibitem{yang2014learn}
J.~Yang, Z.~Lei, and S.~Z. Li, ``Learn convolutional neural network for face
  anti-spoofing,'' \emph{arXiv}, 2014.

\bibitem{Liu2018Learning}
Y.~Liu, A.~Jourabloo, and X.~Liu, ``Learning deep models for face
  anti-spoofing: Binary or auxiliary supervision,'' in \emph{CVPR}, 2018.

\bibitem{Jourabloo2018Face}
A.~Jourabloo, Y.~Liu, and X.~Liu, ``Face de-spoofing: Anti-spoofing via noise
  modeling,'' \emph{arXiv}, 2018.

\bibitem{Krizhevsky2012ImageNet}
A.~Krizhevsky, I.~Sutskever, and G.~Hinton, ``Imagenet classification with deep
  convolutional neural networks,'' in \emph{NIPS}, 2012.

\bibitem{Chingovska2012On}
I.~Chingovska, A.~Anjos, and S.~Marcel, ``On the effectiveness of local binary
  patterns in face anti-spoofing,'' in \emph{Biometrics Special Interest
  Group}, 2012.

\bibitem{Zhang2012A}
Z.~Zhang, J.~Yan, S.~Liu, Z.~Lei, D.~Yi, and S.~Z. Li, ``A face antispoofing
  database with diverse attacks,'' in \emph{ICB}, 2012.

\bibitem{Boulkenafet2017OULU}
Z.~Boulkenafet, J.~Komulainen, L.~Li, X.~Feng, and A.~Hadid, ``Oulu-npu: A
  mobile face presentation attack database with real-world variations,'' in
  \emph{FG}, 2017.

\bibitem{DBLP:conf/cvpr/abs-1812-00408}
S.~Zhang, X.~Wang, A.~Liu, C.~Zhao, J.~Wan, S.~Escalera, H.~Shi, Z.~Wang, and
  S.~Z. Li, ``A dataset and benchmark for large-scale multi-modal face
  anti-spoofing,'' in \emph{CVPR}, 2019.

\bibitem{li2020casiasurf}
A.~Liu, Z.~Tan, X.~Li, J.~Wan, S.~Escalera, G.~Guo, and S.~Z. Li, ``Casia-surf
  cefa: A benchmark for multi-modal cross-ethnicity face anti-spoofing,'' 2020.

\bibitem{race_bias_fr}
``Are face recognition systems accurate? depends on your race,'' 2016,
  \url{https://www.technologyreview.com/s/601786}.

\bibitem{AlviTurning}
M.~Alvi, A.~Zisserman, and C.~Nellaker, ``Turning a blind eye: Explicit removal
  of biases and variation from deep neural network embeddings.''

\bibitem{Wang_2019_ICCV}
M.~Wang, W.~Deng, J.~Hu, X.~Tao, and Y.~Huang, ``Racial faces in the wild:
  Reducing racial bias by information maximization adaptation network,'' in
  \emph{ICCV}, October 2019.

\bibitem{chakka2011competition}
M.~M. Chakka, A.~Anjos, S.~Marcel, R.~Tronci, D.~Muntoni, G.~Fadda, M.~Pili,
  N.~Sirena, G.~Murgia, M.~Ristori \emph{et~al.}, ``Competition on counter
  measures to 2-d facial spoofing attacks,'' in \emph{IJCB}, 2011.

\bibitem{chingovska20132nd}
I.~Chingovska, J.~Yang, Z.~Lei, D.~Yi, S.~Z. Li, O.~Kahm, C.~Glaser, N.~Damer,
  A.~Kuijper, A.~Nouak \emph{et~al.}, ``The 2nd competition on counter measures
  to 2d face spoofing attacks,'' in \emph{2013 International Conference on
  Biometrics (ICB)}.\hskip 1em plus 0.5em minus 0.4em\relax IEEE, 2013, pp.
  1--6.

\bibitem{boulkenafet2017competition}
Z.~Boulkenafet, J.~Komulainen, Z.~Akhtar, A.~Benlamoudi, D.~Samai, S.~E.
  Bekhouche, A.~Ouafi, F.~Dornaika, A.~Taleb-Ahmed, L.~Qin \emph{et~al.}, ``A
  competition on generalized software-based face presentation attack detection
  in mobile scenarios,'' in \emph{2017 IEEE International Joint Conference on
  Biometrics (IJCB)}.\hskip 1em plus 0.5em minus 0.4em\relax IEEE, 2017, pp.
  688--696.

\bibitem{ChallengeCVPR2019}
S.~E. H. J. E. Z. T. Q. Y. K. W. C. L. G. G. I. G. S. Z.~L. Ajian~Liu, Jun~Wan,
  ``Multi-modal face anti-spoofing attack detection challenge at cvpr2019,'' in
  \emph{CVPRW}, 2019.

\bibitem{yu2020searching}
Z.~Yu, C.~Zhao, Z.~Wang, Y.~Qin, Z.~Su, X.~Li, F.~Zhou, and G.~Zhao,
  ``Searching central difference convolutional networks for face
  anti-spoofing,'' \emph{CVPR}, 2020.

\bibitem{wang2020deep}
Z.~Wang, Z.~Yu, C.~Zhao, X.~Zhu, Y.~Qin, Q.~Zhou, F.~Zhou, and Z.~Lei, ``Deep
  spatial gradient and temporal depth learning for face anti-spoofing,''
  \emph{CVPR}, 2020.

\bibitem{he2016deep}
K.~He, X.~Zhang, S.~Ren, and J.~Sun, ``Deep residual learning for image
  recognition,'' in \emph{CVPR}, 2016.

\bibitem{wang2017ordered}
J.~Wang, A.~Cherian, and F.~Porikli, ``Ordered pooling of optical flow
  sequences for action recognition,'' in \emph{WACV}.\hskip 1em plus 0.5em
  minus 0.4em\relax IEEE, 2017, pp. 168--176.

\bibitem{fernando2017rank}
B.~Fernando, E.~Gavves, J.~Oramas, A.~Ghodrati, and T.~Tuytelaars, ``Rank
  pooling for action recognition,'' \emph{TPAMI}, vol.~39, no.~4, pp. 773--787,
  2017.

\bibitem{Boulkenafet2016Face}
Z.~Boulkenafet, J.~Komulainen, and A.~Hadid, ``Face spoofing detection using
  colour texture analysis,'' \emph{TIFS}, 2016.

\bibitem{Li2004Live}
J.~Li, Y.~Wang, T.~Tan, and A.~K. Jain, ``Live face detection based on the
  analysis of fourier spectra,'' \emph{BTHI}, 2004.

\bibitem{patel2016cross}
K.~Patel, H.~Han, and A.~K. Jain, ``Cross-database face antispoofing with
  robust feature representation,'' in \emph{CCBR}, 2016.

\bibitem{HaraCan}
K.~Hara, H.~Kataoka, and Y.~Satoh, ``Can spatiotemporal 3d cnns retrace the
  history of 2d cnns and imagenet?'' 2018.

\bibitem{shen2019facebagnet}
T.~Shen, Y.~Huang, and Z.~Tong, ``Facebagnet: Bag-of-local-features model for
  multi-modal face anti-spoofing,'' in \emph{PRCVW}, 2019, pp. 0--0.

\bibitem{hu2018senet}
J.~Hu, L.~Shen, and G.~Sun, ``Squeeze-and-excitation networks,'' in
  \emph{CVPR}, 2018.

\bibitem{parkin2019recognizing}
A.~Parkin and O.~Grinchuk, ``Recognizing multi-modal face spoofing with face
  recognition networks,'' in \emph{PRCVW}, 2019, pp. 0--0.

\bibitem{qin2020learning}
Y.~Qin, C.~Zhao, X.~Zhu, Z.~Wang, Z.~Yu, T.~Fu, F.~Zhou, J.~Shi, and Z.~Lei,
  ``Learning meta model for zero- and few-shot face anti-spoofing.''
  \emph{Association for Advancement of Artificial Intelligence (AAAI)}, 2020.

\bibitem{DBLP:journals/corr/abs-1907-04047}
\BIBentryALTinterwordspacing
A.~George and S.~Marcel, ``Deep pixel-wise binary supervision for face
  presentation attack detection,'' \emph{CoRR}, vol. abs/1907.04047, 2019.
  [Online]. Available: \url{http://arxiv.org/abs/1907.04047}
\BIBentrySTDinterwordspacing

\bibitem{zhang2019casiasurf}
S.~Zhang, A.~Liu, J.~Wan, Y.~Liang, G.~Guo, S.~Escalera, H.~J. Escalante, and
  S.~Z. Li, ``Casia-surf: A large-scale multi-modal benchmark for face
  anti-spoofing,'' \emph{arXiv:1908.10654}, 2019.

\bibitem{Yi2007Face}
D.~Yi, R.~Liu, R.~F. Chu, Z.~Lei, and S.~Z. Li, \emph{Face Matching Between
  Near Infrared and Visible Light Images}, 2007.

\bibitem{Lei2009Coupled}
Z.~Lei and S.~Li, ``Coupled spectral regression for matching heterogeneous
  faces,'' \emph{CVPR}, pp. 1123--1128, 2009.

\bibitem{lezama2016afraid}
J.~Lezama, Q.~Qiu, and G.~Sapiro, ``Not afraid of the dark: Nir-vis face
  recognition via cross-spectral hallucination and low-rank embedding,'' 2016.

\end{thebibliography}
\vfill\pagebreak
\end{document}